\documentclass[5p,times]{elsarticle}

\usepackage{amssymb}
\usepackage[fleqn]{amsmath, mathtools}
\usepackage{setspace}
\usepackage{amsmath}
\usepackage{soul}
\usepackage{color}
\usepackage{xcolor}
\usepackage{adjustbox}
\usepackage{booktabs}
\usepackage{hyperref}
\usepackage{multirow}
\usepackage{caption}
\usepackage{subcaption}
\usepackage[shortlabels]{enumitem}
\usepackage{makecell}
\usepackage[linesnumbered,ruled,vlined]{algorithm2e}
\usepackage{lineno}
\usepackage{dblfloatfix}
 \usepackage{wrapfig}
\usepackage{graphicx}
\usepackage
	{todonotes}

\journal{Computer Networks}
\begin{document}
\begin{frontmatter}

\title{Federated Learning for 5G Base Station Traffic Forecasting}
\hypersetup{pdfauthor={Vasileios Perifanis, Nikolaos Pavlidis, Remous-Aris Koutsiamanis and Pavlos S. Efraimidis}}

\author[inst1]{Vasileios Perifanis\corref{cor1}}
\ead{vperifan@ee.duth.gr}

\author[inst1]{Nikolaos Pavlidis}
\ead{npavlidi@ee.duth.gr}

\author[inst2]{Remous-Aris Koutsiamanis}
\ead{remous-aris.koutsiamanis@imt-atlantique.fr}

\author[inst1,inst3]{Pavlos S. Efraimidis}
\ead{pefraimi@ee.duth.gr}

\affiliation[inst1]{organization={Democritus University of Thrace},
            addressline={Kimmeria}, 
            city={Xanthi},
            postcode={67100}, 
            country={Greece}}
\cortext[cor1]{Corresponding author}

\affiliation[inst2]{organization={IMT Atlantique -- Inria -- LS2N},
            addressline={4, rue Alfred Kastler}, 
            city={Nantes},
            postcode={44307}, 
            country={France}}
            
\affiliation[inst3]{organization={Athena Research Center},
            addressline={Kimmeria}, 
            city={Xanthi},
            postcode={67100}, 
            country={Greece}}
\begin{abstract}
Cellular traffic prediction is of great importance on the path of enabling 5G mobile networks to perform intelligent and efficient infrastructure planning and management. However, available data are limited to base station logging information. Hence, training methods for generating high-quality predictions that can generalize to new observations across diverse parties are in demand. Traditional approaches require collecting measurements from multiple base stations, transmitting them to a central entity and conducting machine learning operations using the acquire data. The dissemination of local observations raises concerns regarding confidentiality and performance, which impede the applicability of machine learning techniques. Although various distributed learning methods have been proposed to address this issue, their application to traffic prediction remains highly unexplored. In this work, we investigate the efficacy of federated learning applied to raw base station LTE data for time-series forecasting. We evaluate one-step predictions using five different neural network architectures trained with a federated setting on non-identically distributed data. Our results show that the learning architectures adapted to the federated setting yield equivalent prediction error to the centralized setting. In addition, preprocessing techniques on base stations enhance forecasting accuracy, while advanced federated aggregators do not surpass simpler approaches. Simulations considering the environmental impact suggest that federated learning holds the potential for reducing carbon emissions and energy consumption. Finally, we consider a large-scale scenario with synthetic data and demonstrate that federated learning reduces the computational and communication costs compared to centralized settings.
\end{abstract}

\begin{keyword}
Data Privacy \sep Federated Learning \sep Mobile Networks \sep Non-iid data \sep Traffic Forecasting
\end{keyword}

\end{frontmatter}

\section{Introduction} 
\label{sec:intro}
With the development and increasing deployment of fifth generation (5G) cellular networks, networking infrastructures face challenges in handling traffic from heterogeneous devices, ensuring load balancing and maintaining the traffic management reliability \cite{benzaid20205g, kaur_2022_federation}. To improve the services and resource management, accurate traffic forecasting techniques with low prediction error are essential. For example, forecasting Radio Access Network (RAN) slicing demand \cite{Pu20225G} and high-volume traffic in long-term horizons \cite{Pimpinella2022Traffic} are crucial for 5G and beyond networks. Accurate predictions produced by machine learning algorithms can subsequently be used for proper network planning and optimization.

Recent approaches in time-series forecasting primarily involve Multi-Layer Perceptrons (MLPs), Recurrent Neural Network (RNN) models \cite{trinh2020mobile}, such as Long-Short Term Memory (LSTM) and Gated Recurrent Unit (GRU) and Convolutional Neural Networks (CNNs). These models have demonstrated higher predictive accuracy than traditional methods like AutoRegressive Integrated Moving Average (ARIMA) \cite{trinh2018mobile}. 
\begin{figure*}[htbp]
    \centering
    \includegraphics[width=\textwidth]{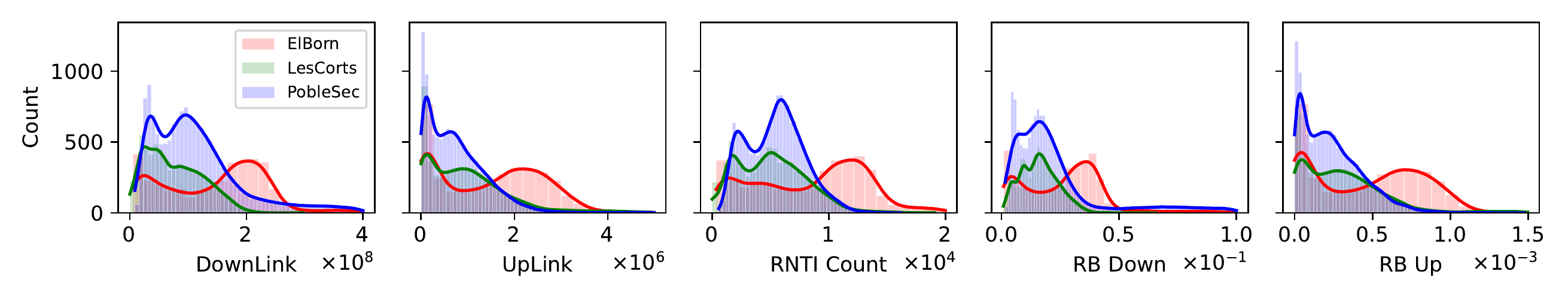}
    \caption{Target values distribution per base station.}
    \label{fig:distributions}
\end{figure*}
\begin{figure}[htbp]
    \centering
     \includegraphics[width=\columnwidth]{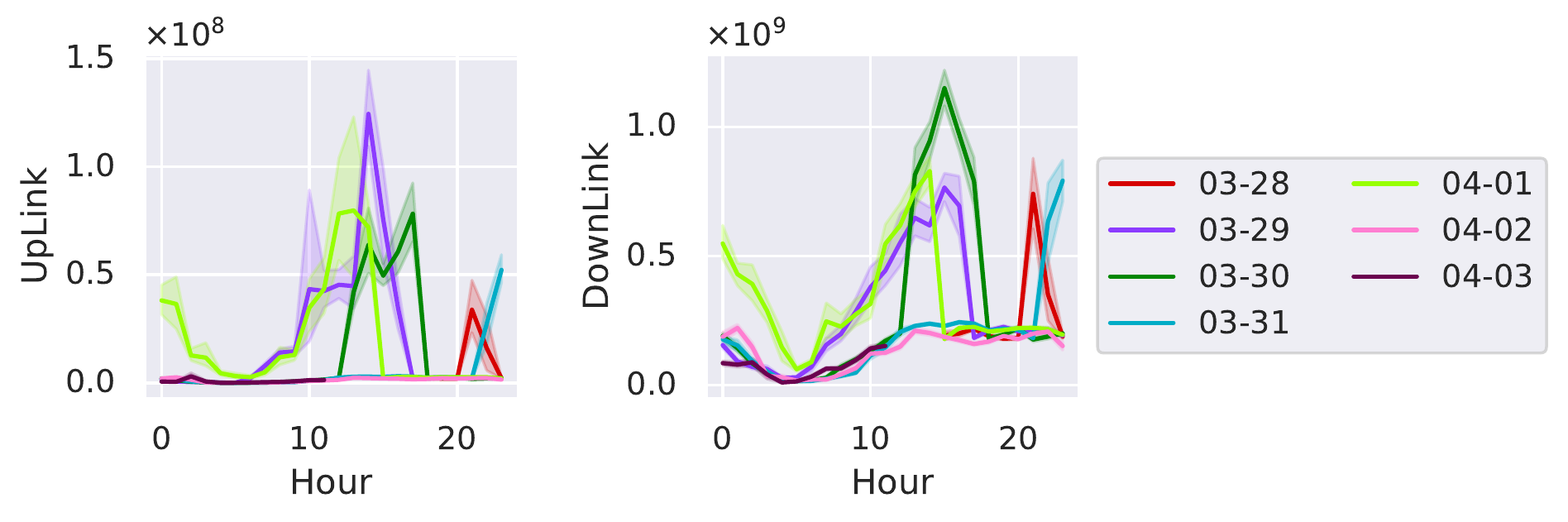}
    \caption{Uplink and downlink distribution per day and hour in ElBorn.}
    \label{fig:changing_distribution}
\end{figure}

Despite the recent advancements in neural networks, their predictive accuracy is limited by the number of observations or available data focus on single-party observations. The collection of data from different operators in 5G and beyond networks can induce storage and communication costs due to the rapid increase of data volume generated by devices, which further leads to scalability issues with respect to the predictive model \cite{Pu20225G}. In addition, regulations and laws worldwide, such as the GDPR, fragment data transmission and model training \cite{goldsteen2022data} and in many cases, business confidentiality issues prevent organizations from sharing their data, hindering real-world applications of traffic demand forecasting.

To address these limitations, federated learning is proposed as a distributed machine learning approach that allows each party to collaborate in a model's training process without exchanging or revealing their own data \cite{mcmahan2017fl}. After collaborative training, the final global model has the potential to generalize to the parties that participated in the distributed training and federated learning holds the promise of reducing communication and computation efforts and also offers a high privacy level when user-based information has to be processed \cite{Pu20225G}.

In the context of \textit{Federated Traffic Prediction for 5G and Beyond Challenge} (our team won the first place in the \href{https://supercom.cttc.es/index.php/ai-challenge-2022}{Federated Traffic Prediction for 5G and Beyond Challenge} {and ranked fourth in the \href{https://aiforgood.itu.int/about-ai-for-good/aiml-in-5g-challenge/}{ITU AI/ML in 5G Challenge 2022}}), we are provided with a dataset from three individual base stations in Spain: \textit{ElBorn}, \textit{LesCorts} and \textit{PobleSec}. The dataset comprises multivariate time-series with variables such as the Radio Network Temporary Identifiers (RNTI) count and the number of allocated resource blocks (RB). The challenge aims to train a model using a federated learning approach to predict the value of five distinct measurements for the next timestep, given as input the measurements from the preceding ten timesteps. Among the five measurements, the challenge focuses on the uplink and downlink transport block sizes, which are crucial for determining the volume and direction of flows.

Federated learning differs significantly from centralized settings and has its own limitations and characteristics. Besides differences in the training process between the two approaches, federated settings pose several challenges regarding data distribution and quantity among participants, leading to accuracy degradation due to weight divergence during training \cite{li2022noniid, rey_2022_fl, zhu2021survey, ma2022survey}. Figure \ref{fig:distributions} presents an example of distribution skew among the three base stations considered in this work regarding the five values to be predicted, which reflect the attribute and target skewness. To quantify the difference among distributions, we performed the Kolmogorov-Smirnov test \cite{kolmogorov_1951} and observed p-values of zero or close to zero (e.g., $1 \times 10^{-190}$) between base stations, suggesting significant differences.

Besides data and target distribution skewness, the corresponding attributes hold temporal characteristics. Since we deal with time-series representations, federated learning poses additional challenges, namely, the temporal and quantity skew \cite{zhu2021survey}. First, the observations are collected in different time periods, i.e., March/April, 2018 - January, 2019 - February/March, 2018 per base station, respectively. Second, the data distributions change over time. An example of a changing distribution in the ElBorn base station for the uplink and downlink measurements is given in Fig. \ref{fig:changing_distribution}. It is observed that (at least) three out of seven days follow different patterns regarding the uplink and downlink measurements, which indicates local temporal skew. In this sense, the pattern from the previous day(s) may not follow the pattern for subsequent days.

Motivated by the success of neural networks applied to centralized settings for time-series forecasting and the potential of federated learning, we study state-of-the-art learning models, including RNNs and a two-dimensional CNN applied to the federated setting. 
One of the most crucial steps in federated learning is the aggregation function, particularly when dealing with non-iid data. Most of the proposed aggregators have been evaluated on different domain tasks such as image classification, sentiment analysis and next character/word prediction \cite{caldas2018leaf}. Thus, in addition to the learning models, we assess the influence of different federated aggregation functions, including Federated Averaging with proximal term (FedProx) \cite{Li2020fedprox}, Federated Averaging with Server Momentum (FedAvgM) \cite{tzu2019fedavgm}, Federated Normalized Averaging (FedNova) \cite{Wang2020fednova} and Adaptive Federated Optimization (FedOpt) and its variations \cite{reddi2021adaptive} in a real-world time-series forecasting task with non-iid data. Finally, we delve into the scalability of federated learning with respect to the training times, the size that needs to be transferred between the server and clients and the predictive accuracy and compare it with the centralized setting.

The main contributions of this paper are summarized as follows:
\begin{itemize}
    \item We identify the challenge of training high quality federated forecasting models under three perspectives: i) data distribution skew, ii) quantity skew and iii) temporal skew.
    \item We experimentally compare five machine learning models trained under three different settings: i) individual learning, where each base station performs local training and evaluation, ii) centralized learning and iii) federated learning.
    \item We conduct extensive experiments using nine different aggregation algorithms, some of which are specifically designed for handling non-iid data.
    \item We show that preprocessing techniques can significantly reduce the prediction error and have greater impact than state-of-the-art aggregation algorithms for non-iid data.
    \item We open source a framework for evaluating federated time-series forecasting\footnote{\href{https://github.com/vperifan/Federated-Time-Series-Forecasting}{https://github.com/vperifan/Federated-Time-Series-Forecasting}}, which is flexible, fully customizable and can serve as a baseline for future federated time-series research.
\end{itemize}

Through our experiments we show that federated models achieve comparable predictive accuracy to individual and centralized learning. Additionally, we demonstrate that local preprocessing techniques, specifically outlier flooring and capping can lead to huge error reduction, while state-of-the-art aggregation algorithms for non-iid data do not outperform simpler approaches. Finally, we show that federated learning can scale up to numerous clients offering lower computational demands, training times and communication costs than centralized learning. To the best of our knowledge, this is the first study that applies federated learning to raw LTE data, considers attribute and target distribution, quantity and temporal skew and evaluates the prediction error by combining five machine learning models and nine aggregation functions. 

The rest of this work is structured as follows. Section \ref{sec:related} introduces the main concepts of time-series forecasting and federated learning and reviews related work on federated time series forecasting. Section \ref{sec:methodology} defines the federated traffic prediction problem and presents the design of the federated setting as well as the federated aggregation functions. Section \ref{sec:experiments} discusses the experimental results. Finally, Section \ref{sec:conclusion} concludes our work.

\section{Preliminaries and Related Work}
\label{sec:related}
\subsection{Time-Series Forecasting}
Time-series forecasting is an essential task in a wide range of real-life problems, such as weather prediction\cite{rasp2020weather}, energy consumption \cite{alhussein2020load}, sales \cite{loureiro2018sales} and traffic prediction \cite{trinh2018mobile}. In recent years, there has been a substantial increase in machine learning methods applied to time-series data. Unlike classic methods such as ARIMA-based models, deep learning approaches have shown great success in modeling time-series due to their ability to map non-linear correlations. Neural architectures for modeling time-series data can be classified into three categories \cite{benitez2021review}: (i) Fully Connected Neural Networks such as MLPs, ii) Recurrent Networks such as RNNs, LSTMs and GRUs and iii) Convolution Networks such as CNNs.

In this work, we apply federated time-series analysis on raw LTE data to model the temporal dependence of traffic demand. Traffic forecasting is a vital task in the 5G era, given the rapid traffic growth and evolving patterns over time \cite{iwamura20155g}. In the context of traffic prediction, Trinh et al. \cite{trinh2018mobile} compared LSTM to MLP and ARIMA and showed that LSTM significantly outperforms the other models. Feng et al.\cite{feng2018deeptp} proposed DeepTP, an LSTM-based network which models the spatial dependence of time-series and temporal changes, demonstrating that the proposed framework surpasses traditional methods such as ARIMA and LSTM without enhancements. Chen et al. \cite{chen2018cran} introduced a clustered LSTM-based model for multivariate time-series modeling. Sebastian et al. \cite{sebastian2020traffic} suggested a decomposition of time-series into seasonal, trend and cycle components, feeding the resulting vectors into the GRU model. Gao \cite{gao20225g} proposed an enhancement of an LSTM network with stationarity evaluation (SLSTM), specifically designed for 5G network measurements. Pimpinella et al. \cite{Pimpinella2022Traffic} focused on predicting long-term traffic peaks in LTE measurements and compared clustered LSTM to Additive Decomposition and Seasonal ARIMA. Finally, in \cite{Bejaro2021Social}, the authors focused on predicting traffic series by integrating exogenous data such as scheduled social events.

Although the recent advancements in modeling time-series data, most works focus on centralized learning, which differs significantly from federated learning. In addition, federated time-series forecasting has yet to be explored in-depth, particularly in the context of mobile networks. In this paper, we examine five different model architectures that span the three model categories presented in \cite{benitez2021review}. Although Transformer and AutoEncoder-based models have shown promising results in time-series forecasting \cite{lim2021tft, Nguyen_Quanz_2021}, they are not considered in this study due to their high computational and energy consumption costs. In federated learning, where communication and resource efficiency are critical, the chosen models aim to minimize these costs while maintaining adequate predictive performance. To our knowledge, this is the first study that provides insights using different time-series modeling under federated learning. In addition, although long-term cluster-based training has demonstrated great potential \cite{Pimpinella2022Traffic}, in this work we focus on federated training for short-term forecasting and leave clustered-based training for future work. Finally, we plan to further integrate exogenous data such as weather information or social events \cite{Bejaro2021Social} to further benefit the forecasting accuracy.

\subsection{Federated Learning}
Federated Learning is a machine learning technique, originally proposed by McMahan et al. \cite{mcmahan2017fl}, which enables collaborative training of data across multiple devices. This approach promises to revolutionize the model training process as it differs from traditional centralized machine learning techniques where all local data are uploaded to a central server, as well as from more conventional decentralized approaches which often assume that local data samples to be identically distributed \cite{kairouz2021fl}. The federated setting has been evaluated in numerous machine learning tasks, including image classification \cite{mcmahan2017fl}, next word prediction \cite{caldas2018leaf}, anomaly detection \cite{pei_2022_anomaly}, collaborative filtering \cite{perifanis2022fedncf} and time-series forecasting \cite{taik2020load}. 

\begin{figure}[t!]
    \centering
    \includegraphics[width=\columnwidth]{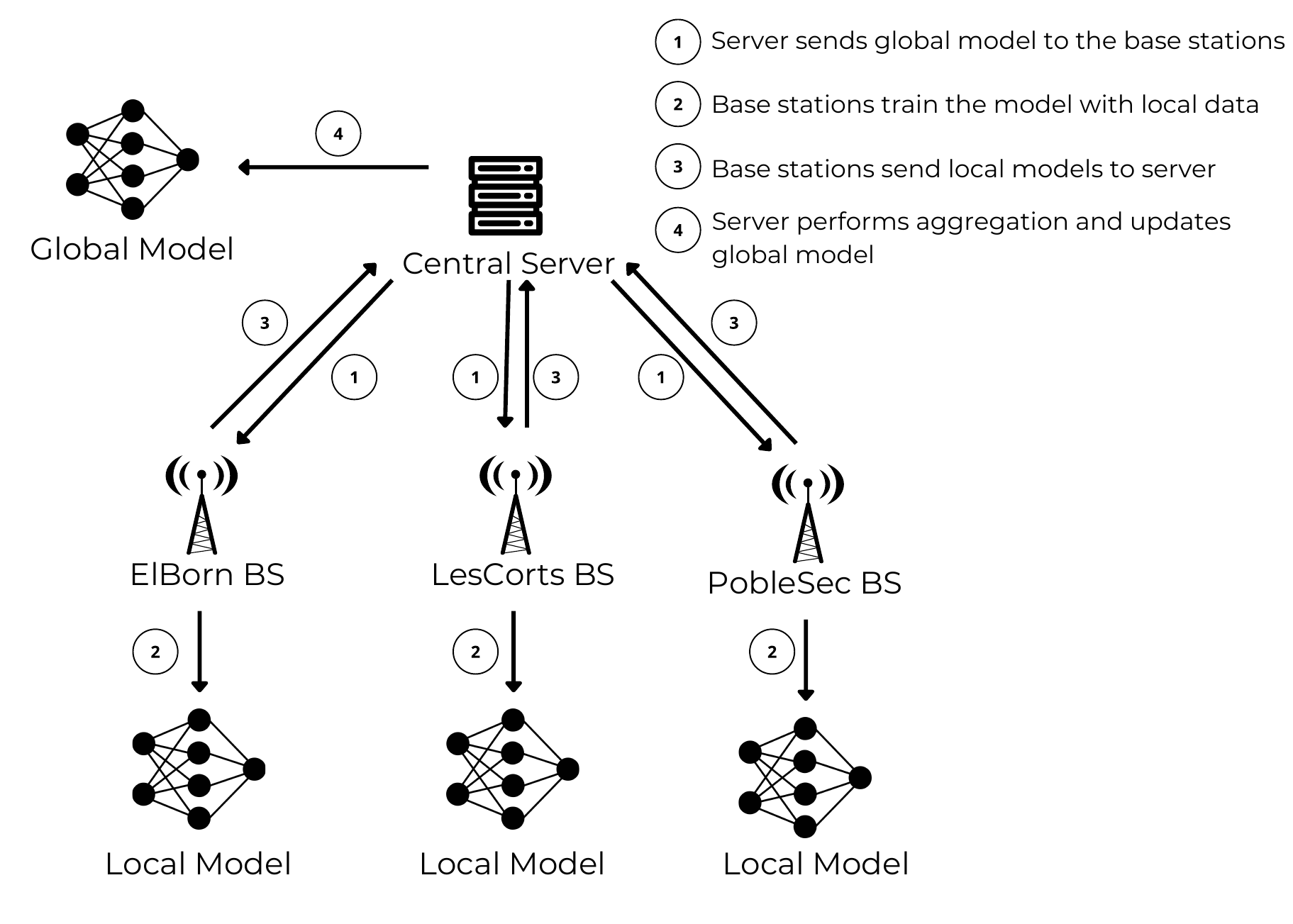} 
    \caption{Federated learning process.}
    \label{fig:fl_overview}
\end{figure}

During the federated learning process, each participating entity performs local training on its private data using the model weights shared by a central server. Then, in contrast to traditional machine learning, participants only share their calculated model parameters. The central server gathers these parameter updates and aggregates them using an aggregation function. The aggregation result corresponds to the new global model, which is then shared to the participants for the next round. An overview of one round of the federated process is given in Fig. \ref{fig:fl_overview}. In this paper, we follow the common approach of federated training on raw base station time-series data, i.e., a third-party is responsible for collecting weight updates per federated round, summarizing the results using an aggregation algorithm and distributing the aggregated data to participants for local training and weight updating.

\subsubsection{Federated Learning on non-iid Data}
While federated learning offers high-quality model generation, extensive research has been conducted on the impact of non-iid data on model accuracy \cite{li2022noniid, zhu2021survey, kairouz2021fl}. In particular, base station observations hold most categories of non-iid data, i.e., attribute, label, quantity and temporal skew \cite{zhu2021survey, kairouz2021fl}. In our case, as we use historical data to provide one-step predictions, the attribute and label values are presented in both training features and target values. Assuming a local distribution of $P(x_i, y_i) = P(x_i | y_i)P(y_i)$, feature and label distribution skew implies that $P(x_i)$ and $P(y_i)$ differ among participating entities, respectively. Quantity skew indicates that the number of observations varies among participants. Lastly, temporal skew implies that the timesteps of observations can differ or the feature distribution can be changing over time (Fig. \ref{fig:changing_distribution}). Consequently, we study the effectiveness of federated learning by considering a combination of non-iid data categories, which, to our knowledge has yet to be explored. The most closely related work regarding the evaluation of federated learning under non-iid data is presented in Li et al. \cite{li2022noniid}, in which non-iid categories are considered independently on image and tabular datasets for the classification task.

\subsection{Federated Time-Series Forecasting}
Time-series forecasting has a wide range of applications in various industries and lots of practical applications. To comply with regulations and reduce computation and communication overhead, federated learning has also been explored in the context of time-series prediction.

Taik et al. \cite{taik2020load} employed a federated LSTM model for the household load forecasting task. After generating the global model, participants perform a fine-tuning step using their local data to achieve personalization. Liu et al. \cite{liu2020traffic} proposed FedGRU with the FedAvg algorithm \cite{mcmahan2017fl} to predict the number of vehicles in an area. Before initiating federated training, the participating entities form clusters using the K-Means algorithm. After grouping, participants train a global model for each cluster. Liu et al. \cite{liu2021anomaly} introduced a federated attention-based CNN-LSTM model for anomaly detection using time-series data with the FedAvg algorithm. In addition, the communication cost of federated training is minimized using a selection-based gradient compression algorithm. Briggs et al. \cite{briggs2022fl} proposed FedLSTM applied to load forecasting similar to \cite{taik2020load}. FedLSTM integrates exogenous data such as weather information and applies a clustering step among participating entities by considering the weight similarity of local models. Fekri et al. \cite{Fekri2022Fl} adapted LSTM to the federated setting and used FedAvg and FedSGD \cite{mcmahan2017fl} for the load forecasting task, similar to \cite{briggs2022fl}. Before training, the data owners perform local preprocessing, i.e., participants execute feature scaling locally. In this work, we argue that local scaling leads to inconsistencies during training and we show that global scaling results in lower forecasting error. Similarly, \cite{Pu20225G} used a FedLSTM to predict traffic load for inter-slice resource orchestration, considering a differentiated loss for overestimated and underestimated model predictions.

All of the previous works in federated time-series forecasting do not raise the issue of non-iid data and evaluate the generated global models only using the FedAvg algorithm. In this work, we raise the challenge of training federated models under the presence of four categories of non-iid data, evaluate different aggregation functions and show that local preprocessing techniques can lead to higher predictive accuracy. In addition, similar works only adapt a single neural network architecture to the federated setting, e.g., LSTM \cite{taik2020load, briggs2022fl, Fekri2022Fl, Pu20225G} or GRU \cite{liu2020traffic}. Hence, insights regarding different federated models applied to time-series forecasting have yet to be provided. We emphasize that in our use case, the participating entities (base stations) are limited to three and therefore, clustered federated learning \cite{liu2020traffic, briggs2022fl} cannot be directly applied. Finally, since federated learning requires several rounds of operations, many studies review the environmental impact of federated learning regarding energy consumption and $CO_{2eq}$ emissions \cite{guerra2023cost}. In our experiments we measure the communication cost regarding the total uplink and downlink transmission of model weights and the energy consumption for training the global model with respect to the carbon footprint ($CO_{2eq}$). 

\section{Methodology}
\label{sec:methodology}
\subsection{Problem Formulation}
We first describe the problem regarding \textit{individual learning}, where a single party holds observations and performs local training. Let $X_{t} = \{x_{\{t,1\}}, ..., x_{\{t, d\}}\}$ represent the measurements at timestep $t$, with $d$ being the number of variates. For a given $t$, we consider a window of past observations $T \in [t-T+1, t]$ and $X_{t}^* = \{X_{t-T+1}, ..., X_t\}$. The objective is to use the past observations $X_{t}^*$ to predict the measurements for the next step, $\hat{y}_{t+1}$. By utilizing the entire dataset of measurements $D_{ind} = \bigcup_{i \in m} X_{\{t,i\}}^*$, where $m$ is the number of multivariate time series containing the window $T$, the goal is to build a model that can generalize on unseen future series.

In \textit{centralized learning}, the measurements are transmitted to a third-party who trains a machine learning model using the combined data. Given the combined data from $n$ participants $D_{cen} = \bigcup_{i \in n} D_{ind}^i$, the goal is to build a model that can generalize to future series, at least for the $n$ participants. 

\textit{Federated learning} can be seen as an intermediate setting between individual and centralized learning. More precisely, $n$ participants collaborate to build a forecasting model that can generalize to their future observations. Each participant $p \in n$ holds their own time-series and trains a learning model locally for a limited number of epochs. Then, participants transmit their locally-learned weights to an aggregator, which generates an averaged model. This process continues until the global model generalizes to the observations of the $n$ participants. Consequently, in federated learning, each participating entity optimizes a model based on the local data, similar individual learning. Meanwhile, the generated global model has the capability of generalization to $n$ entities, which is the ultimate goal of machine learning.

\subsection{Federated Traffic Forecasting Design}
\begin{figure}[t!]
    \centering
    \includegraphics[width=\columnwidth]{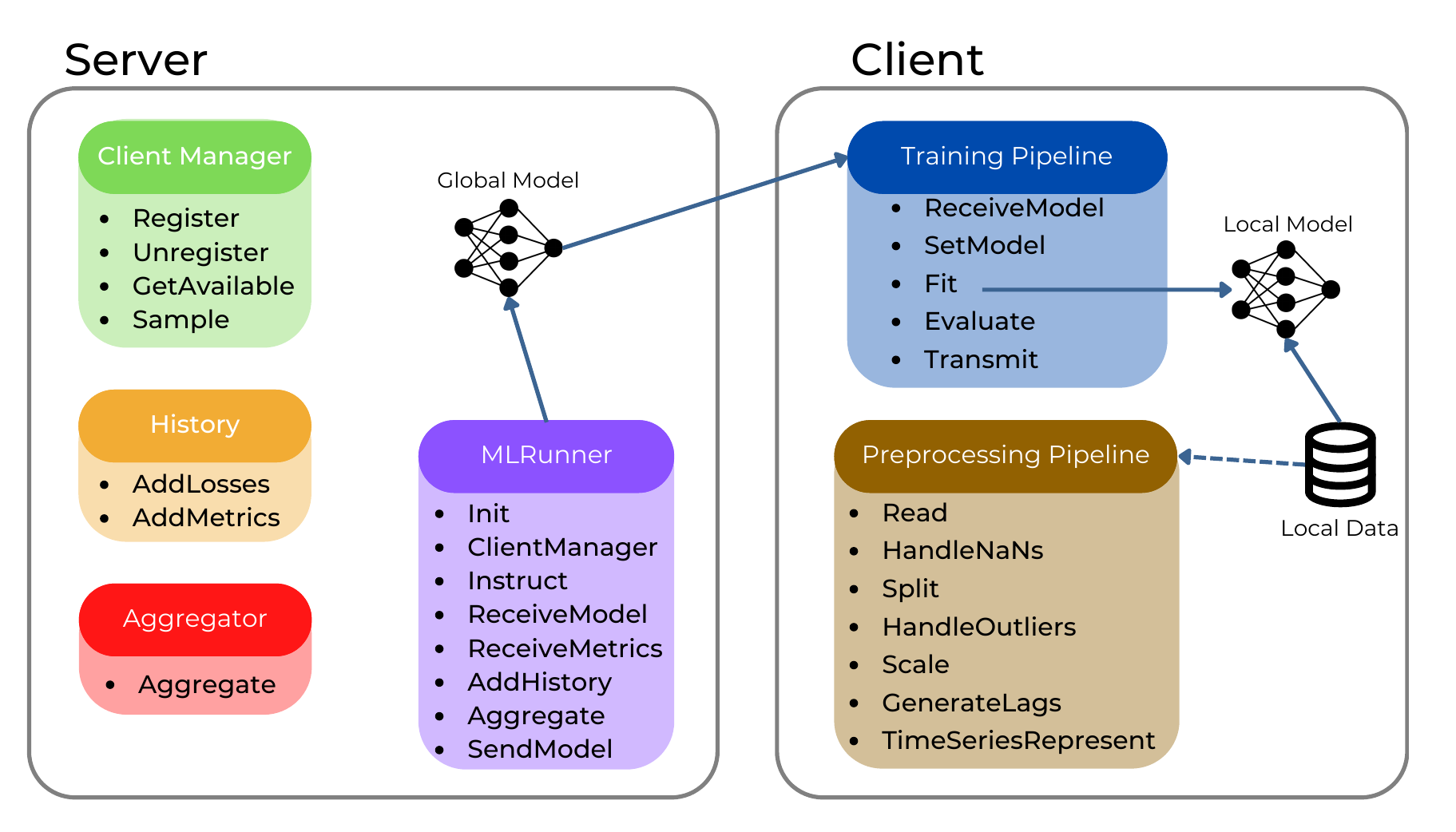} 
    \caption{Federated time-series forecasting overview.}
    \label{fig:fl_components}
\end{figure}

To simulate a realistic traffic forecasting scenario, we have developed a customizable framework that can be extended for generalized time-series forecasting. Our architecture is based on the Flower federated learning framework \cite{Beutel2020Flower}, uses PyTorch \cite{Paszke2019Pytorch} as the backbone deep learning library and state-of-the-art aggregation algorithms are integrated using the NumPy package \cite{Harris2020Numpy}. Our code can be easily extended with additional PyTorch models or NumPy-based aggregation algorithms. Note that the transition from federated learning to the individual or centralized setting is also supported. Figure \ref{fig:fl_components} illustrates the main components of the federated pipeline. It consists of two core entities, the \textit{Server} and the \textit{Client}. The Server is responsible for selecting available clients per federated round, logging historical information and aggregating model parameters. A Client represents individual entities that participate in the federated computation, which perform the preprocessing and training pipelines using the local data. The following subsections describe the main components of our approach.

\subsubsection{Data Preprocessing}
In general, time-series data are collected automatically from sensors or logging mechanisms, such as base stations logging network measurements or homes equipped with smart meters for measuring electricity consumption. It is natural for the data to hold redundant information, errors or unusual spikes, making preprocessing decisive for the subsequent learning task. Our preprocessing pipeline consists of four main steps:

\paragraph{Data Cleansing} This step handles missing or corrupted data and identifies and manages outliers. We adopt a simple technique of transforming missing values to zeroes and using the flooring and capping technique for managing outliers. Zero transformation is selected over simple removal to preserve the continuous data. In addition, imputing missing values with a constant may not be representative for time-series data, and estimating them can be time and energy consuming. 
\paragraph{Data Split} The data are divided into three subsets for model training, evaluation and testing. Specifically, the data per base station are split into 60\% for training, 20\% for validation and 20\% for testing. 
\paragraph{Scaling} We perform Min-Max normalization to eliminate the influence of value ranges.
\paragraph{Time-series Representation} The final step generates a dataset of sliding windows governed by the window size $T$.

Our preprocessing pipeline is designed in such a way to handle the three different learning settings. More precisely, after data collection, the owner feeds the observations to the preprocessing pipeline. Then, a handler for missing values is executed and the dataset is split to train and validation sets. After splitting, an outlier handler is applied to the training set. We emphasize that outliers should not be transformed on the validation and test sets since it concerns samples that are not provided during training. This way, we can also target estimating future peaks \cite{Pimpinella2022Traffic} without information loss. After outlier handling, the features are being scaled and finally, the data are represented as time-series using a sliding window of $T$.

It is worth noting that preprocessing techniques heavily influence the model's performance. However, there has been limited research for this stage applied to federated learning, i.e., how different entities should handle and transform their data. A similar observation is made in \cite{Fekri2022Fl}, where the authors used local preprocessing and scaled the data individually on each client. We argue that individual preprocessing leads to inconsistent transformations among participants and we experimentally show that global scaling leads to higher model quality. In addition, we experimentally show that the application of flooring and capping for outliers handling leads to a remarkable forecasting error decrease.

\subsubsection{Federated Training}
\begin{figure}[t!]
    \centering
    \includegraphics[width=\columnwidth]{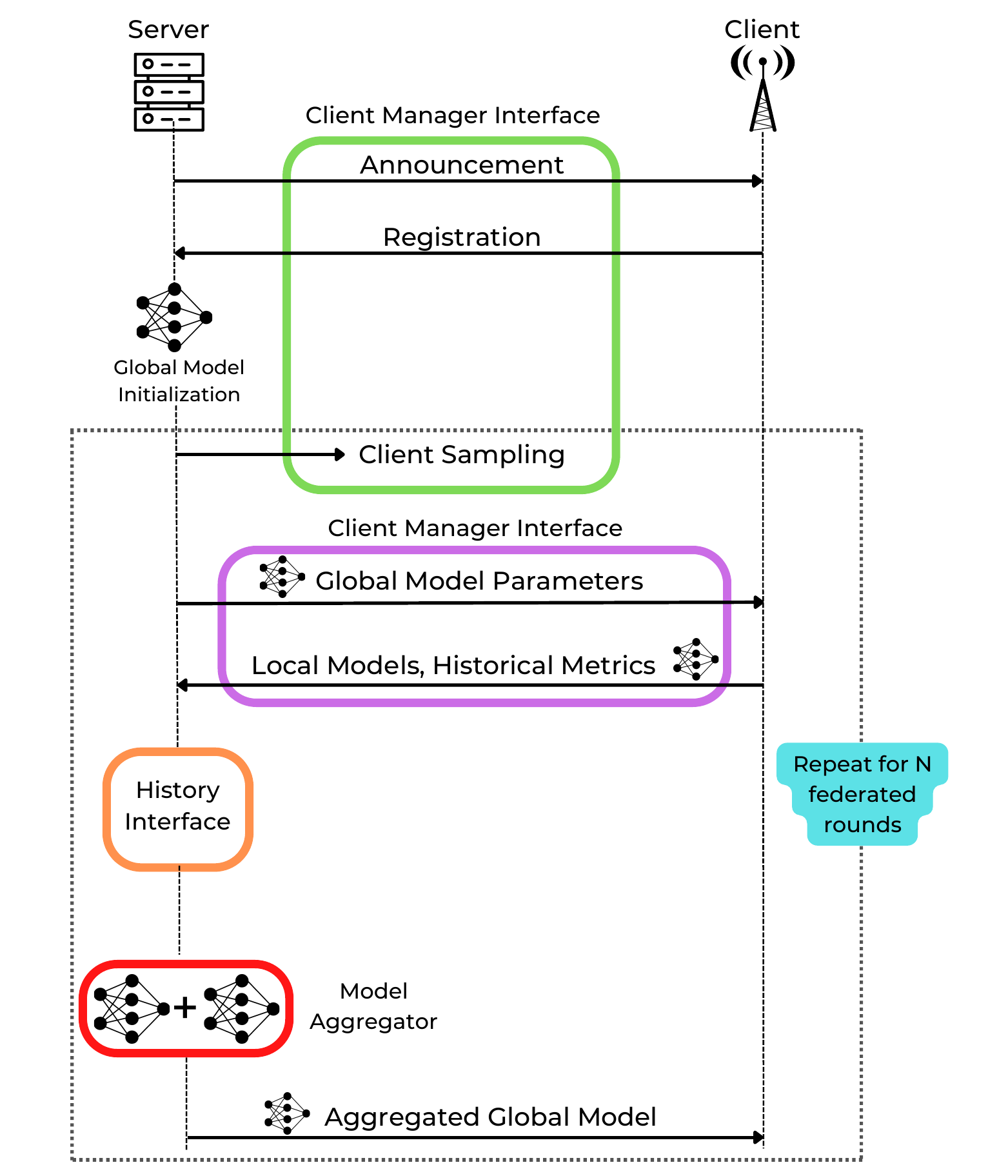} 
    \caption{Server -- Client operations during federated training.}
    \label{fig:fed_training}
\end{figure}

Initially, the server announces the computation and clients, such as base stations, register as participants. Sampled clients receive the current global model from the server and perform local training using their data. After local training, the updated model is transmitted back to the server along with historical information such as the loss and evaluation metrics. The server aggregates the received local models, updates the global model and the process repeats for multiple federated rounds. Fig. \ref{fig:fed_training} illustrates the operations and communication between the server and clients. 

Upon completion of federated training, the server transmits the final global model to participants. By design, federated learning can be executed dynamically at different time intervals to capture new observations and improve the predictive accuracy. Moreover, participants can also conduct a local fine-tuning step to drift the model closer to their local data, thereby providing higher quality forecasting predictions. Ultimately, to generate predictions on unseen data, each client should apply the preprocessing operations during the announcement phase and feed the transformed data to the trained model. The predictions should then be transformed back to the original feature range to obtain the final prediction.
 
\subsubsection{Federated Aggregation}
\label{aggregators}
\SetKwInput{KwInput}{Input}                
\SetKwInput{KwOutput}{Output}              

\begin{algorithm}[ht!]
\small
\KwInput{local datasets $D^i$, number of clients $N$, number of federated rounds $T$, number of local epochs $E$, learning rate $\eta$, $beta$, $\beta_1, \beta_2 \in [0,1)$ for FedAvgM, FedYogi \& FedAdam, $\lambda$ degree of adaptivity}
\KwOutput{Final global model parameters vector $w^T$}

\textbf{Server executes:}

initialize $w^0$

\For{t=0,1,...,T-1}    
{ 
    Sample a set of parties $S_t$
    
    $n \gets \sum_{i \in S_t}|D^i|$

    \For{$i \in S_t$  \textbf{\emph{in parallel}}}
    {
    send the global model $w^t$ to client $C_i$

    $\Delta w_i^t , \textcolor{orange}{\tau_i}  \gets $ \textbf{LocalTraining}($i, w^t$)
    }

    $\Delta W \gets \sum_{i \in S_t} \frac{|D^i|}{n} \Delta w_k^t$
    
    For FedAvg/FedProx:

    $w^{t+1} \gets w^t$ - $\eta$ $\Delta W$

    \colorbox{orange}{For FedNova:}

    $w^{t+1} \gets w^t$ - $\eta$ \textcolor{orange}{$\frac{\sum_{i \in S_t}|D^i|\tau_i}{n}$} $\sum_{i \in S_t} \frac{|D^i|}{n\textcolor{orange}{\tau_i}} \Delta w_k^t$

    \colorbox{magenta}{For FedAvgM:}

    \textcolor{magenta}{$u_t \gets \beta u_{t-1} + \Delta W$}

    $w^{t+1} \gets w^t$ - $u_t$

    \colorbox{olive}{For FedAdagrad:}

    \textcolor{olive}{$u_t \gets u_{t-1} + \Delta W^2$}

    $w^{t+1} \gets w^t$ + $\eta$ $\frac{\Delta W}{\textcolor{olive}{\sqrt{u_t} + \lambda}}$

    \colorbox{teal}{For FedYogi:}
    
    \textcolor{teal}{$m_t \gets \beta_1 m_{t-1} + (1-\beta_1) \Delta W$}
    
    \textcolor{teal}{$u_t \gets u_{t-1} - (1-\beta_2) \Delta W^2 sign(u_{t-1} - \Delta W^2)$}

    $w^{t+1} \gets w^t$ + $\eta$ \textcolor{teal}{$\frac{m_t}{\sqrt{u_t} + \lambda}$}

    \colorbox{brown}{For FedAdam:}

    \textcolor{brown}{$m_t \gets \beta_1 m_{t-1} + (1-\beta_1) \Delta W$}

    \textcolor{brown}{$u_t \gets \beta_2 u_{t-1} + (1-\beta_2) \Delta W^2$}

    $w^{t+1} \gets w^t$ + $\eta$ \textcolor{brown}{$\frac{m_t}{\sqrt{u_t} + \lambda}$}
}

return $w^T$

\textbf{Client executes:}

For every algorithm: $L(w;\textbf{b}) = \sum_{(x,y) \in \textbf{b}} l(w;x;y)$

For FedProx: $L(w;\textbf{b}) = \sum_{(x,y) \in \textbf{b}} l(w;x;y) \textcolor{red}{+ \frac{\mu}{2}{||w-w^t||}^2}$

\caption{FedAvg Algorithm and it's variations \colorbox{red}{FedProx}, \colorbox{orange}{FedNova}, \colorbox{magenta}{FedAvgM}, \colorbox{olive}{FedAdagrad}, \colorbox{teal}{FedYogi}, \colorbox{brown}{FedAdam}}
\label{alg:aggregators}
\end{algorithm}
One of the most important steps in federated learning is model aggregation. In this phase, the central server collects and aggregates models from participants to update the global model's state. This process poses difficulties, especially when dealing with non-iid and heterogeneous data among the clients. For this reason, numerous research efforts have focused on aggregation algorithms. 

The most common algorithm used in federated learning is \textbf{FedAvg}, proposed by McMahan et. al \cite{mcmahan2017fl}. The aggregation is performed by computing a weighted average of client models based on the data quantity, giving more influence to clients with more samples. However, in some cases, FedAvg can lead to objective inconsistency, where the global model converges to a stationary point of a mismatched objective function which can be significantly different from the true (global) objective. This inconsistency is the result of non-iid and heterogeneous data presence. To mitigate this issue, alternatives to FedAvg have emerged to better capture data heterogeneity.

To provide theoretical guarantees under non-iid data, Li et al. proposed a re-parametrization of FedAvg called \textbf{FedProx} \cite{Li2020fedprox}. \textbf{FedProx} is a generalization of the FedAvg algorithm that offers robust convergence, particularly in highly heterogeneous settings. More precisely, 
the algorithm introduces a tunable term $\mu$ to control the local objective, which limits the distance between the previous and current model weights. 

Another algorithm is \textbf{FedNova} \cite{Wang2020fednova}, which normalizes local model updates during averaging. FedNova's primary principle is that it averages the normalized local gradients by dividing them with the number of local steps that each client performed individually, as opposed to averaging the cumulative local gradient without any normalization step.

\textbf{FedAvgM} \cite{tzu2019fedavgm} is a FedAvg variation that leverages server momentum. Momentum is computed by iteratively multiplying previous model updates with a hyper-parameter $\beta$ in each epoch while simultaneously adding the new updates. 

Finally, an attempt to create federated versions of adaptive optimizers, namely \textbf{FedAdagrad}, \textbf{FedYogi} and \textbf{FedAdam}, is presented in \cite{reddi2021adaptive}. These methods aim to address the heterogeneous data, improve performance and reduce communication costs. However, their efficiency depends on hyper-parameter ($\lambda, \beta_1, \beta_2$) optimization. Algorithm \ref{alg:aggregators} summarizes the state-of-the-art aggregation methods considered in this work.

\section{Experiments}
\label{sec:experiments}
\subsection{Dataset Description}
We were provided with a dataset of real LTE Physical Downlink Control Channel (PDCCH) measurements collected from three different base stations in Barcelona, Spain. The dataset is made available under the Federated Traffic Prediction for 5G and Beyond Challenge and a preliminary work using the corresponding data is presented in \cite{trinh2018mobile}. More specifically, the dataset comprises three base stations that are treated as separate clients:
\begin{enumerate}
    \item \textbf{LOCATION 1, ``ELBORN'':} 5421 samples, collected from 2018-03-28 15:56:00 to 2018-04-04 22:36:00
    \item \textbf{LOCATION 2, ``LESCORTS'': } 8615 samples, collected from 2019-01-12 17:12:00 to 2019-01-24 16:20:00
    \item \textbf{LOCATION 3, ``POBLESEC'':} 19909 samples, collected from 2018-02-05 23:40:00 to 2018-03-05 15:16:00
\end{enumerate}
\begin{figure}[htbp]
    \centering
    \includegraphics[height=0.96\textheight]{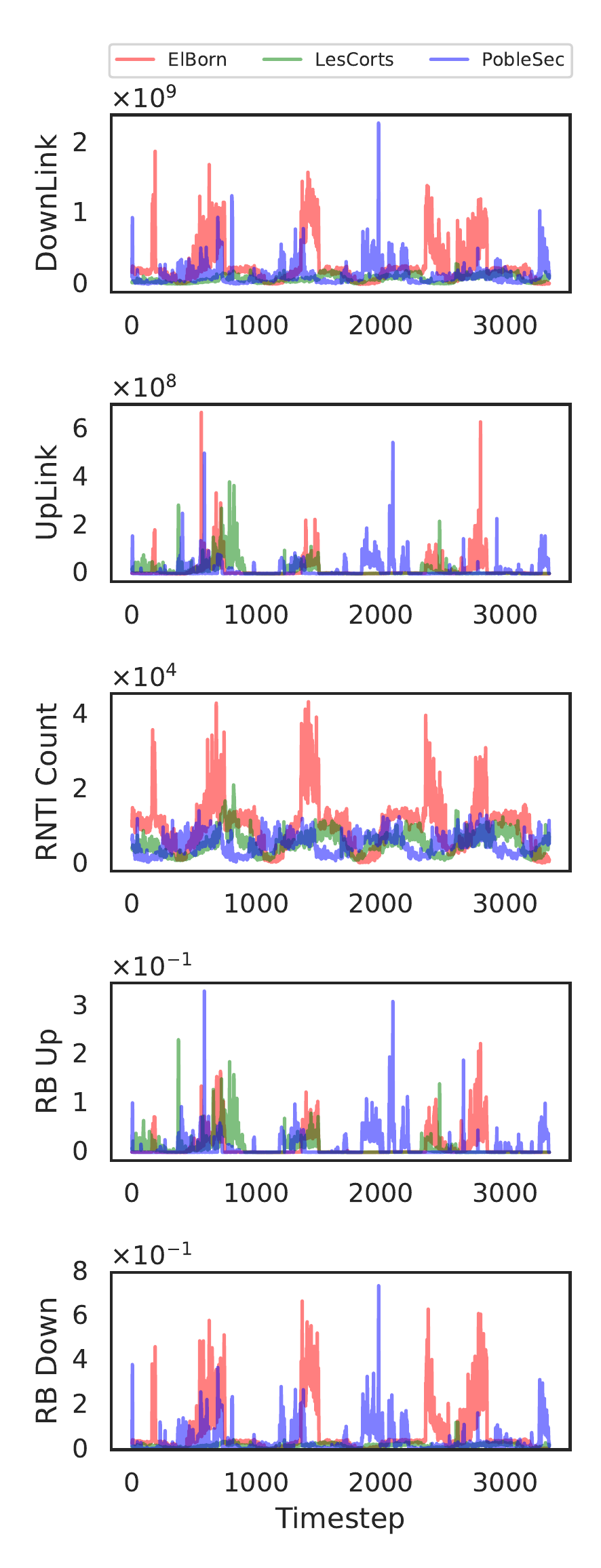}
    \caption{Target series per base station.}
    \label{fig:series}
\end{figure}

For each site, eleven features are provided related to downlink and uplink traffic. Specifically, at each timestep the measurements concern the \textbf{DownLink} and \textbf{UpLink} traffic, the number of RNTIs (\textbf{RNTI Count}), the number of allocated resource blocks (\textbf{RB Up} and \textbf{RB Down)} and their variances (\textbf{RB Up Var} and \textbf{RB Down Var}), the Modulation and Coding Schemes (MCS) (\textbf{MCS Up} and \textbf{MCS Down}) and their variances (\textbf{MCS Up Var} and \textbf{MCS Down Var}). Each timestep represents an aggregation of these measurements into a two-minute interval. The objective is to predict the first five measurements for the next timestep using as input the observations with a window of $T=10$ per base station. Fig. \ref{fig:series} illustrates the five input series per base station that serve as target values. Note that the samples for LesCorts and PobleSec are trimmed to match the number of training instances on ElBorn. For completeness, Fig. \ref{fig:input_series2} shows the remaining input series per base station.

\subsection{Metrics}
We evaluate the models performance using the \textbf{Mean Absolute Error (MAE)}, \textbf{Root Mean Squared Error (RMSE)} and \textbf{Normalized RMSE (NRMSE)}. MAE measures the absolute difference between the expected and predicted output. RMSE measures how far predictions are from the ground truth values using the Euclidean distance. We use MAE and RMSE to report the obtained results by considering all five values that are being predicted. Since the values for UpLink and DownLink are very large in scale (on the order of billions of bits), we use the NRMSE to facilitate the comparison between experiments for the UpLink and DownLink measurements. These traffic types are crucial as they affect the overall network performance, including latency, throughput and reliability. The accurate prediction of future uplink and downlink traffic enables efficient resource allocation, leading to improved network performance. The considered metrics are defined as follows:

\begin{equation}
    MAE = \frac{1}{n}\sum_{i=1}^{n}|\hat{y_i}-y_i|,
\end{equation}
\begin{equation}
    RMSE = \sqrt{\frac{\sum_{i=1}^{n}(\hat{y_i}-y_i)^2}{n}},
\end{equation}
\begin{equation}
    NRMSE = \frac{1}{\overline{y}}\sqrt{\frac{\sum_{i=1}^{n}(\hat{y_i}-y_i)^2}{n}}.
\end{equation}
 
In the specified metrics, $n$ is the number of data points, $y_i$ is the $i$-th measurement and $\hat{y_i}$ is its corresponding prediction.

In addition to the metrics used to evaluate the prediction accuracy of our learning models, we measure the energy and environmental cost of the proposed solution with respect to $CO_{2eq}$ emissions and model power consumption using the Carbontracker Python Library \cite{anthony2020carbontracker}.

\subsection{Models and Learning Setting}
\label{learning_settings}
We conducted extensive experiments on the given dataset to investigate the effectiveness of deep learning models applied to the federated setting. We compared the model architectures using \textit{individual learning}, \textit{centralized learning} and \textit{federated learning}. The following architecture were considered:

\paragraph{MLP} It is a simple feed-forward artificial neural network with three hidden layers, $h=\{256, 128, 64\}$.  

\paragraph{RNN} It is a vanilla Recurrent Neural Network with a single layer of 128 units. The output of RNN is then fed to a MLP with one hidden layer and 128 units.

\paragraph{LSTM} It is an improvement over RNN, limiting the problem of exploding gradients and can model longer sequential data \cite{benitez2021review}. Similar to RNN, we select a LSTM layer with 128 units and the output is fed to a MLP with one hidden layer and 128 units.

\paragraph{GRU} Similar to LSTM, the GRU model addresses the problem of exploding gradients on RNNs. It is a simplified version of LSTM with lower computational parameters. It follows the architecture of RNN and LSTM, i.e., a GRU layer with 128 units and a MLP with one layer of 128 units.

\paragraph{CNN} This type of network operates directly on raw data using convolution layers. The selected CNN takes as input a three-dimensional matrix of size $(1, T, \#variates)$ and feeds it to four two-dimensional convolutional layers with filter sizes $\{16, 16,$ $32, 32\}$. The output is forwarded to a two-dimensional average pooling layer and finally to a fully-connected layer of 128 units. 

To ensure fair comparisons, we keep consistent model architectures across the three learning settings. We note that grid search was not performed on the selected architectures due to practical constraints in real-world federated learning scenarios.

For each model and learning setting, we used the Adam optimizer \cite{kingma2015Adam} with a learning rate of $0.001$. The activation function among model layers is ReLU. During training, we optimize the Mean Squared Error (MSE) and use a batch size of $128$. Although other loss function were tested (such as MAE and Focal Loss), MSE yielded similar predictive accuracy.

For individual and centralized learning, we use a maximum of $270$ epochs and for the federated setting we select $30$ rounds with $3$ local epochs per participant. Due to the low number of sites available in the dataset we perform no client sampling. The number of epochs for the individual and centralized setting is selected in such a way to provide fair comparison with federated learning, i.e., in federated learning the complete number of epochs is 270 ($\text{rounds} \times \text{local epochs} \times \#\text{participants}$). The model with the lowest MSE on the validation set during training was selected in all settings. For the individual and centralized approaches, we also integrated an early stopping component with a patience of 50 epochs. 

The experimental study is conducted on a workstation running Ubuntu 20.04 with 128 GB memory and a RTX 3060 GPU. The programming language is Python 3.8. For each experiment, we report the averaged results and the stability using standard deviation after retraining the models from scratch using different seeds.
\setcounter{figure}{7}
\begin{figure*}[hbp!]
    \centering
    \includegraphics[width=\textwidth]{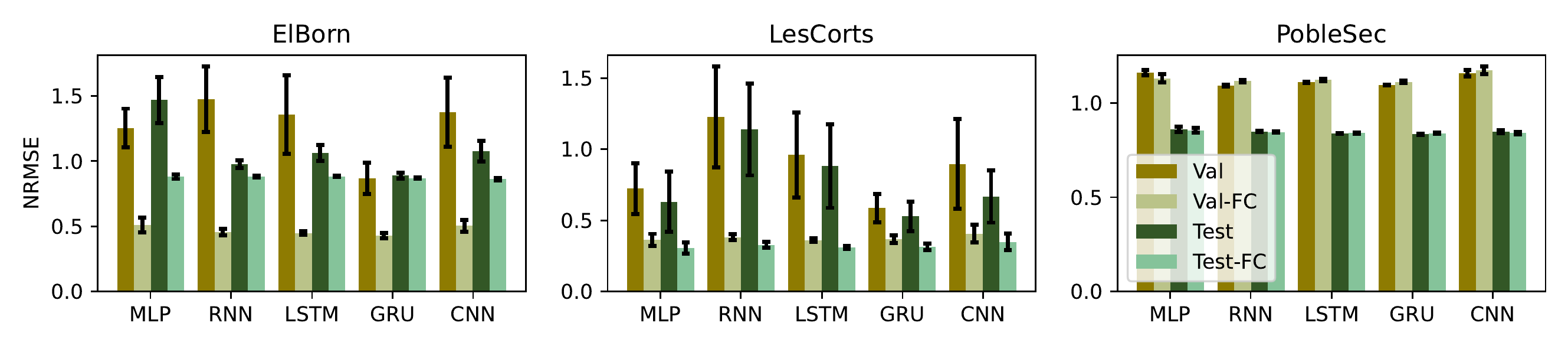}
    \caption{Averaged NRMSE with and without flooring and capping per base station.}
    \label{fig:with_without_bs}
\end{figure*}

\subsection{Preprocessing Stage Influence}
In this experimental evaluation, we study the influence of preprocessing on predictions using the federated setting. Specifically, we investigate the use of outlier transformation with the flooring and capping (FC) technique and the use of global and local scalers. Note that FC is applied only to training series per base station and not on the validation or test sets, to ensure accurate prediction for future series. At this stage we perform federated training using the FedAvg algorithm \cite{mcmahan2017fl}.
\setcounter{figure}{6}
\begin{figure}[t!]
    \centering
    \includegraphics[width=0.8\columnwidth]{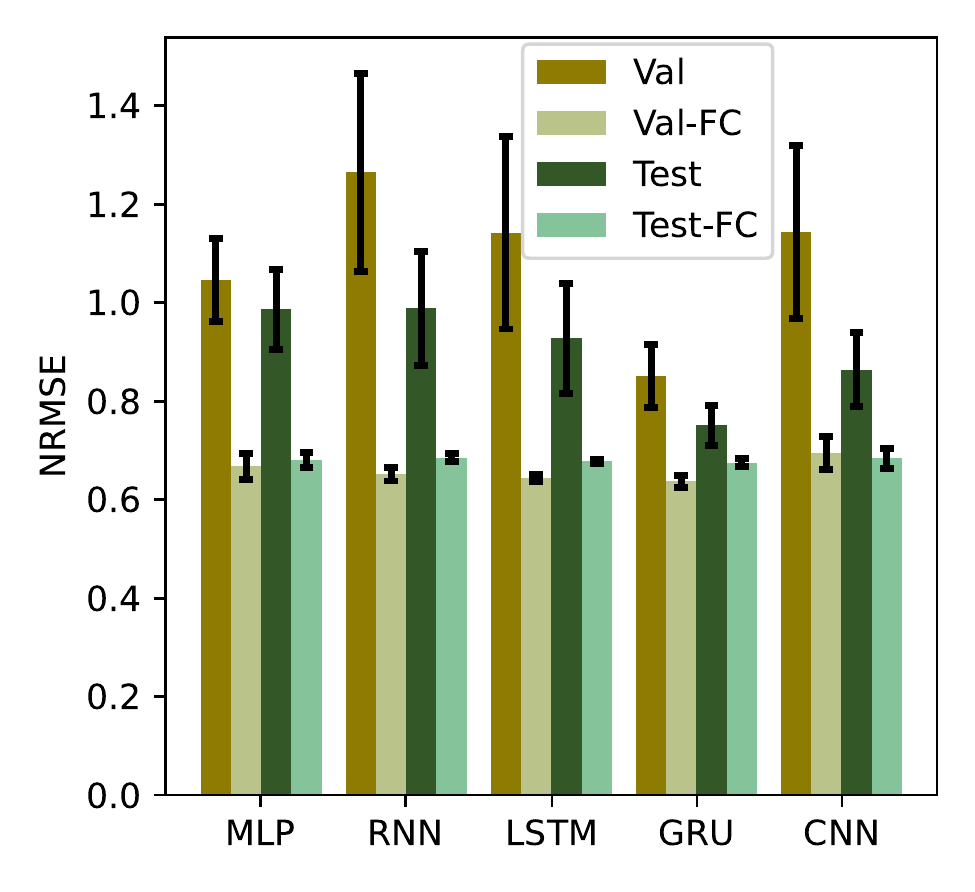} 
    \caption{Averaged NRMSE with and without flooring and capping. FC denotes the utilization of flooring and capping.}
    \label{fig:with_without_outliers}
\end{figure}

We start by comparing models trained with and without FC, where we find that FC consistently results in lower errors compared to training with raw data, regardless of the specified percentiles for flooring and capping. Nevertheless, after a grid search, we found that the optimal percentiles per base statio are (10, 90) for both ElBorn and LesCorts and (5, 95) for PobleSec. Fig. \ref{fig:with_without_outliers} shows the averaged NRMSE with and without FC on the validation and testing sets, demonstrating a remarkable decrease in forecasting error and improved robustness when FC is enabled.

To provide more comprehensive results, we also present the NRMSE per base station, individually. In Fig. \ref{fig:with_without_bs}, we observe that training the models using FC reduces the errors on ElBorn and LesCorts, but does not improve the NRMSE in PobleSec. We attribute this behavior to the random spikes and outliers present the in the later base station.

Another integral step of preprocessing is feature scaling. In federated learning, we can allow each participant to locally compute the transformation or use a global variable to scale the features. In this work, we select Min-Max normalization, which transforms each observation according to the given minimum and maximum value. Fekri et al. \cite{Fekri2022Fl} utilized the first approach, i.e., each participant scales the values using the local observations. However, this transformation leads to different scales per participant, which in turn can lead the model to inconsistencies between the input features and the target variables. Besides, a fair comparison between centralized and federated learning requires scaling using the global minimum and maximum. Fig. \ref{fig:local_global_minmax} illustrates the averaged validation and testing NRMSE with local \cite{Fekri2022Fl} and global scaling, showing that global scaling leads to lower error and standard deviations on both the validation and testing sets.

Based on the previous observations, we continue the experimental study using flooring and capping along with scaling local features using the global minimum and maximum. We emphasize that generating those global values requires a communication between participating entities. However, such computation is lightweight since participants only share two local values, and we select the global scaling approach owing to higher forecasting quality.

\subsection{Learning Setting Comparison}
In this experimental evaluation, we compare the predictive accuracy of the selected models in the three learning settings. We use the same learning parameters for each setting and the FedAvg algorithm in federated learning to ensure a fair comparison. The averaged MAE and RMSE, which consider all five predicted values, are reported in Table \ref{tab:results_rmse_mae}, while the averaged NRMSE, which considers only the UpLink and DownLink values, is reported in Table \ref{tab:results_nrmse}. The highest quality model per learning setting is denoted in bold and the second best is underlined.
\setcounter{figure}{8}
\begin{figure}[t!]
    \centering
    \includegraphics[width=0.8\columnwidth]{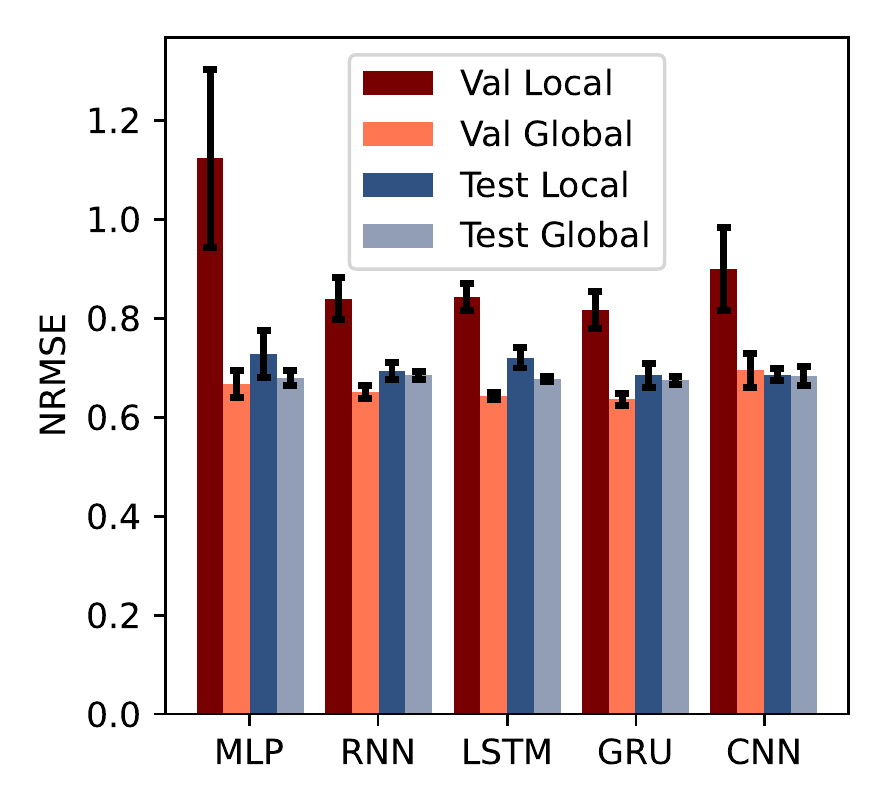} 
    \caption{Averaged NRMSE using local and global scaling.}
    \label{fig:local_global_minmax}
\end{figure}

\begin{table*}[ht!]
\centering
\caption{Averaged RMSE($\times10^7$) and MAE($\times 10^6$) considering all five predicted values on the test set per learning setting.}
\scriptsize
\centering
\begin{tabular}{l|c|c|c|c|c|c}
    & \multicolumn{2}{c|}{Individual} & \multicolumn{2}{c|}{Centralized} & \multicolumn{2}{c}{Federated}  \\
    \hline
    Model & RMSE & MAE    & RMSE & MAE   & RMSE & MAE\\
    \hline
    MLP &  2.8485$\pm$4.2807($\times 10^5$) & 7.4638$\pm$8.8464($\times 10^4$) & 2.8447$\pm$4.6895($\times 10^5$)   & 7.5474$\pm$1.5786($\times 10^5$)    &    2.9655$\pm$5.7121($\times 10^5$) &    7.6694$\pm$1.4936($\times 10^5$)
    \\
    \hline
    RNN & 2.7917$\pm$8.6265($\times 10^4$)   & 7.3339$\pm$3.886($\times 10^4$) & 2.8498$\pm$1.7813($\times 10^5$) & 7.4373$\pm$5.6791($\times 10^4$)   &  2.9042$\pm$2.8667($\times 10^5$) &   7.6031$\pm$8.7780($\times 10^4$)  \\
    \hline
    LSTM & \underline{2.7699}$\pm$1.0200($\times 10^5$) & \underline{7.2944}$\pm$3.6675($\times 10^4$)     &
    2.8392$\pm$9.1586($\times 10^4$)& \underline{7.4018}$\pm$2.8927($\times 10^4$)     &   
     \underline{2.8486}$\pm$1.8695($\times 10^5$) & \underline{7.3632}$\pm$4.2940($\times 10^4$) \\
    \hline
    GRU & \textbf{2.7652}$\pm$6.6683($\times 10^4$)& \textbf{7.231}$\pm$3.4017($\times 10^4$)  & 
    \textbf{2.8103}$\pm$1.2527($\times 10^5$) & \textbf{7.3187}$\pm$4.1427($\times 10^4$)    &    
     \textbf{2.8275}$\pm$1.6222($\times 10^5$) & \textbf{7.3316}$\pm$4.7141($\times 10^4$)   \\
    \hline
    CNN  & 2.8484$\pm$4.3976($\times 10^5$) & 7.5244$\pm$9.6178($\times 10^4$) & \underline{2.8283}$\pm$6.72082($\times 10^5$)  & 7.4774$\pm$2.0889($\times 10^5$) & 2.9702$\pm$4.7189($\times 10^5$)   &     7.5549$\pm$9.6281($\times 10^4$)  \\ 
\midrule
\end{tabular}
\label{tab:results_rmse_mae}
\end{table*}

\begin{table}[ht!]
\centering
\caption{Averaged NRMSE considering the UpLink and DownLink values on the test set per learning setting.}
\scriptsize
\centering
\begin{tabular}{l|c|c|c}
    & {Individual} & {Centralized} & {Federated}  \\
    \hline
    MLP &   0.6755$\pm$0.01204   & \underline{0.6705}$\pm$0.013      &   0.6797$\pm$0.0158
    \\
    \hline
    RNN &  0.6631$\pm$0.0171    & 0.6852$\pm$0.0085   &     0.6856$\pm$0.0081
    \\
    \hline
    LSTM    & \textbf{0.6417}$\pm$0.0095    & 0.6927$\pm$0.0058  &     \underline{0.6776}$\pm$0.0049
    \\
    \hline
    GRU &   \underline{0.6589}$\pm$0.0051    &   0.6734$\pm$0.0057    & \textbf{0.6747}$\pm$0.0079
    \\
    \hline
    CNN    & 0.6687$\pm$0.0151    &   \textbf{0.6701}$\pm$0.0219     &     0.6836$\pm$0.0198
    \\ 
\midrule
\end{tabular}
\label{tab:results_nrmse}
\end{table}

\begin{figure*}[hb!]
    \centering
    \includegraphics[width=0.75\textwidth]{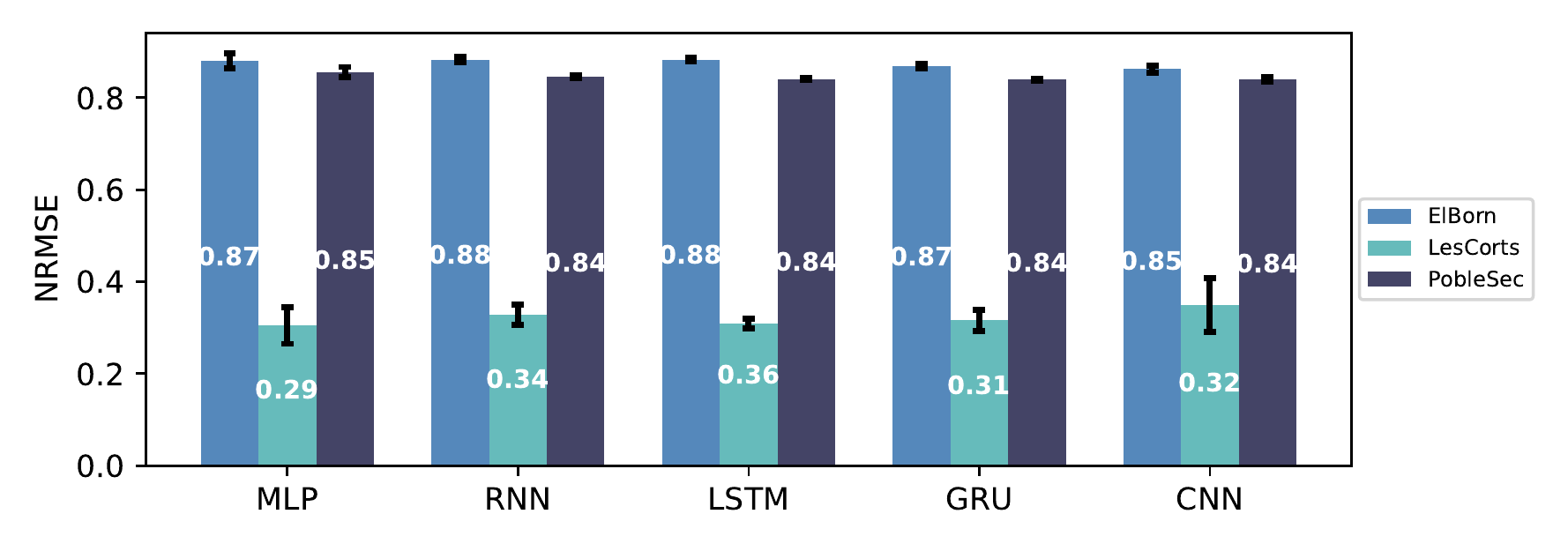}
    \caption{Averaged test NRMSE per model, learning setting and base station.}
    \label{fig:nrmse_per_model_setting}
\end{figure*}

Overall, we observe that all models provide relatively similar forecasting error. In the \textit{individual setting}, the top two models are LSTM and GRU since they present the lowest error in all the metrics. In \textit{centralized learning}, it is difficult to identify the highest quality model(s). However, LSTM and GRU perform the best regarding MAE, GRU and CNN regarding RMSE and MLP and CNN regarding NRMSE. Nonetheless, all models result in similar prediction errors. Zooming into model stability, which provides us with a model robustness quality indicator, LSTM and GRU outperform other models. On the other hand, CNN has the highest standard deviation, indicating that although it can provide high-quality predictions, its generalization ability is susceptible to randomness. Finally, LSTM and GRU outperform the rest of the models in the \textit{federated setting}, resulting also in lower standard deviation. Based on the above observations, LSTM and GRU are the highest quality models in each of the three learning settings. 

\begin{figure*}[ht!]
    \centering
    \includegraphics[width=\textwidth]{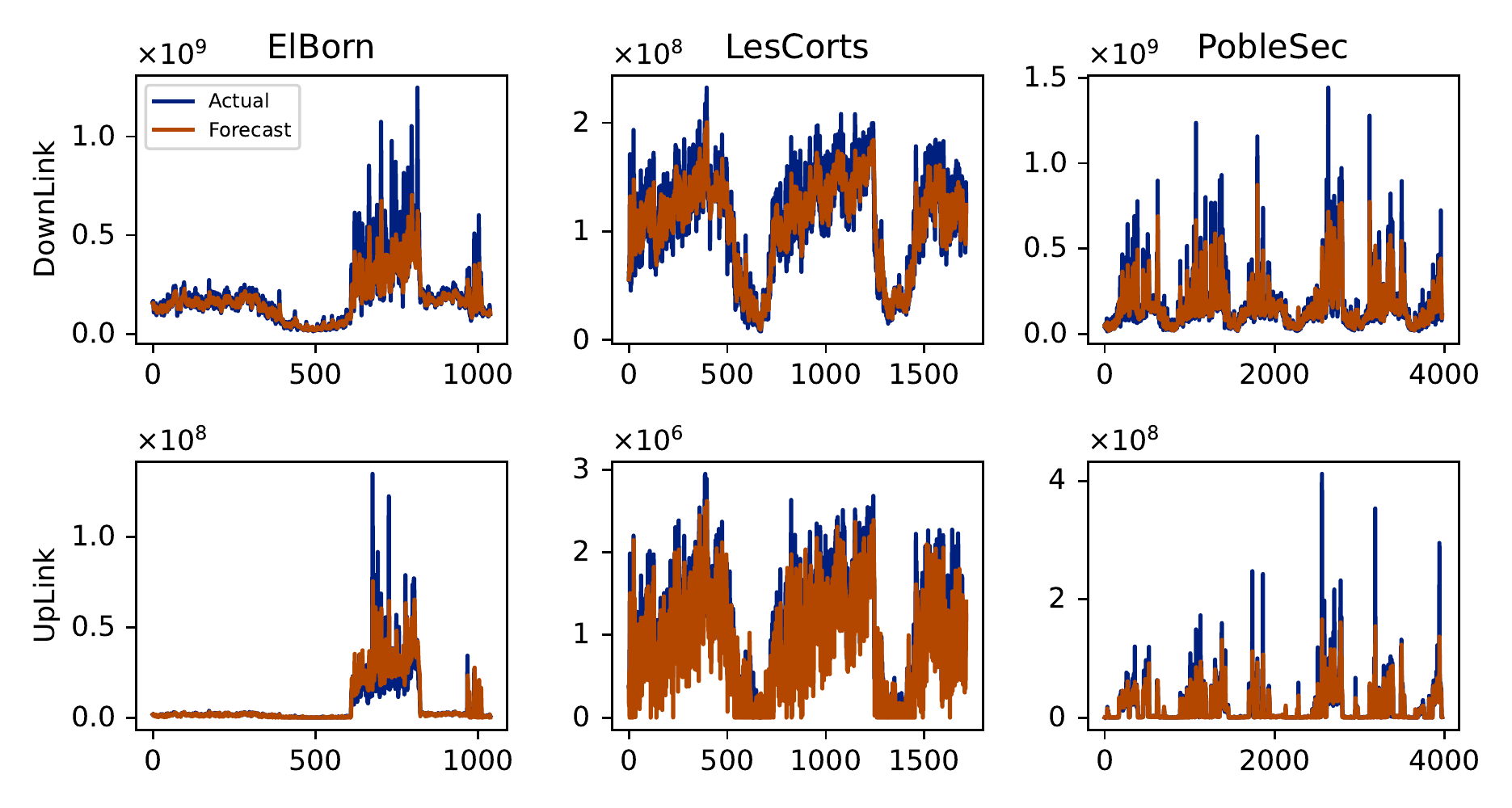}
    \caption{Predictions of the federated LSTM model against ground truth values per base station.}
    \label{fig:forecast}
\end{figure*}
Directly comparing the identified top models among learning settings regarding the prediction accuracy, we end up with the following observation: the individual setting results in the highest quality, followed by centralized and federated learning, which is consistent with previous research in time-series forecasting \cite{briggs2022fl}. In all settings, MLP and CNN have higher instability in both validation and test sets. In contrast, LSTM and GRU provide high robustness on the validation sets, leading to higher quality predictions on the test sets.

In addition to the averaged results, we are interested in the predictive accuracy on each base station, individually. Fig. \ref{fig:nrmse_per_model_setting} illustrates the obtained test NRMSE per site on the federated setting. The forecasting error is almost equivalent per model and learning setting and the predictive accuracy is higher on LesCorts. In addition, the employed flooring and capping technique effectively handles some outliers for the rest measurements such as the uplink values, leading to lower prediction error. On the other hand, the forecasting error on ElBorn and PobleSec is about 75\% higher. This behavior is not only attributed to extreme outliers, but also to higher values for the uplink and downlink measurements. Specifically for the federated setting, we observe that the LSTM model does not only result in high predictive accuracy, but also high stability individually, as well as collectively. Fig. \ref{fig:forecast} shows the forecast of the federated LSTM model per base station regarding the downlink and uplink measurements against the ground truth values. For completeness, Fig. \ref{fig:forecast2}, presents the forecasts for the remaining target values. From the visualization of forecasts, it is evident that the predictions on LesCorts are on par with the ground truth, while on ElBorn and PobleSec, the model fails to capture spikes and extremely high values. Note, however, that in the latter base stations, the respective measurements are much higher than those on LesCorts, i.e., about 90\% and over 97\% higher for the downlink and uplink measurements compared to LesCorts, respectively. Nevertheless, the patterns of measurements are well represented and the predictive accuracy of federated learning is on par with individual and centralized settings.

\begin{figure}[t!]
    \centering
    \includegraphics[width=0.8\columnwidth]{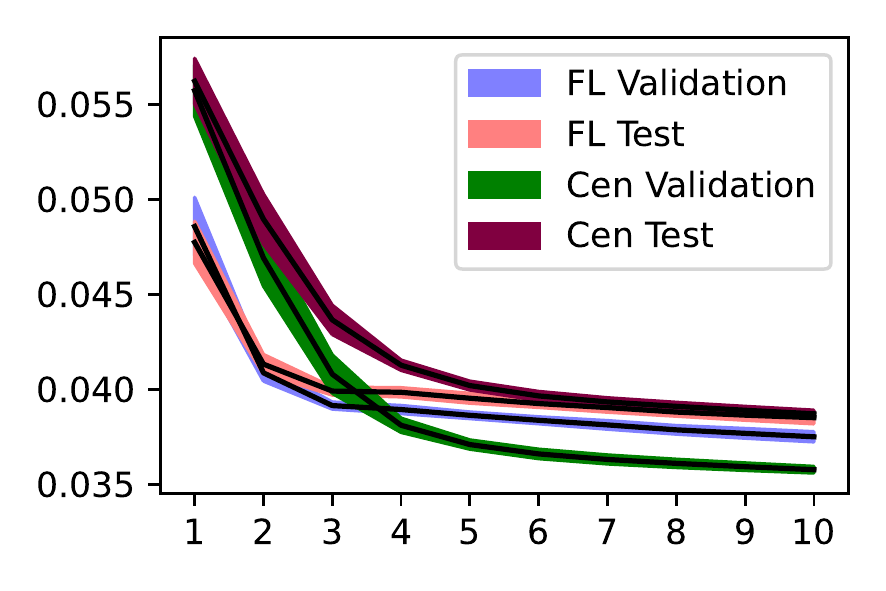}
    \caption{Convergence of centralized and federated settings regarding MAE.}
    \label{fig:cen_fl_curves}
\end{figure}

Fig. \ref{fig:cen_fl_curves} verifies the convergence of federated learning with respect to the centralized setting regarding the MAE on the training and validation set (using the scaled data and considering the averaged MAE). Based on the results obtained from the setting comparison, we argue that federated learning can provide high quality time-series forecasting accuracy with \textit{generalization} ability (that cannot be ensured under individual learning), while overcoming issues related to \textit{business confidentiality}, \textit{regulations} and \textit{collaboration}, unlike centralized learning. 

\subsection{Local Fine-Tuning}
Although individual learning leads to lower forecasting error per base station, it only captures the temporal dynamics of a single base station, thus limiting the generalization ability. In addition, centralized learning requires the transmission of obtained measurements, which limits the utilization of private data. In this subsection, we perform a local fine-tuning similar to \cite{taik2020load, briggs2022fl} to observe whether additional local epochs achieve personalization. Each model is trained from scratch 10 times using different initialization seeds. Table \ref{tab:local_fn} reports the obtained NRMSE per base station as well as the averaged NRMSE per learning setting on the test set. The highest quality model per learning setting considering the averaged NRMSE is denoted with bold. For ElBorn, LesCorts and PobleSec, individually, the top models are colored with orange, olive and red, respectively. Note that in individual learning there is no local fine-tuning step since the model directly learns on base station local data. 

\begin{table*}[ht!]
\centering
\caption{Averaged test NRMSE per model, learning and base station, with and without local fine-tuning. Fine-tuned models are denoted with the "-LF" suffix.}
\begin{adjustbox}{width=\textwidth,center}
\centering
\begin{tabular}{l|c|c|c|c|c|c|c|c|c|c|c|c}
    & \multicolumn{4}{c|}{Individual} & \multicolumn{4}{c|}{Centralized} & \multicolumn{4}{c}{Federated}  \\
    \hline
    Model & ElBorn & LesCorts & Poblesec & Avg & ElBorn & LesCorts & Poblesec & Avg & ElBorn & LesCorts & Poblesec & Avg           \\
    \hline
    MLP & \multirow{2}{*}{0.8361$\pm$0.0119} & \multirow{2}{*}{0.3411$\pm$0.0592} & \multirow{2}{*}{0.8428$\pm$0.0153} & \multirow{2}{*}{0.6733$\pm$0.0222} & 0.8735$\pm$0.0198 & 0.2791$\pm$0.0102 & 0.8499$\pm$0.0241 & 0.6675$\pm$0.0126 & 0.8805$\pm$0.0141 & 0.2909$\pm$0.0281 & 0.8505$\pm$0.0064 & 0.6832$\pm$0.0118 \\
    MLP-LF & & & & & 0.8370$\pm$0.0205 & 0.2671$\pm$0.0119 & 0.8873$\pm$0.0341 & 0.6638$\pm$0.0171 & 0.8257$\pm$0.0125 & 0.2915$\pm$0.0069 & 0.8773$\pm$0.0294 & 0.6698$\pm$0.0132 \\
    \hline
    RNN & \multirow{2}{*}{0.8537$\pm$0.0041} & \multirow{2}{*}{0.3423$\pm$0.0299} & \multirow{2}{*}{0.8382$\pm$0.0019} & \multirow{2}{*}{0.6781$\pm$0.011} & 0.8729$\pm$0.0044 & 0.3445$\pm$0.0246 & 0.8411$\pm$0.0021 & 0.6862$\pm$0.0095 & 0.8785$\pm$0.0091 & 0.3245$\pm$0.0148 & 0.8455$\pm$0.0034 & 0.6828$\pm$0.0071 \\
    RNN-LF & & & & & 0.8622$\pm$0.0049 & 0.3268$\pm$0.0129 & 0.8431$\pm$0.0028 & 0.6774$\pm$0.0051 & 0.8586$\pm$0.0041 & 0.3282$\pm$0.0071 & 0.8411$\pm$0.0028 & 0.6759$\pm$0.0019 \\
    \hline
    LSTM & \multirow{2}{*}{0.8314$\pm$0.0064} & \multirow{2}{*}{0.3292$\pm$0.0332} & \multirow{2}{*}{\textcolor{red}{\textbf{0.8339}}$\pm$0.0012} & \multirow{2}{*}{0.6648$\pm$0.0119} & 0.8763$\pm$0.0044 & 0.3501$\pm$0.0119 & 0.8440$\pm$0.0017 & 0.6901$\pm$0.0051 & 0.8826$\pm$0.0042 & 0.3068$\pm$0.0084 & 0.8389$\pm$0.0016 & 0.6761$\pm$0.0038 \\
    LSTM-LF & & & & & 0.8498$\pm$0.0246 & 0.3033$\pm$0.0134 & 0.8426$\pm$0.0015 & 0.6652$\pm$0.0058 & 0.8603$\pm$0.0413 & 0.2929$\pm$0.0166 &  0.8381$\pm$0.0016 & 0.6638$\pm$0.0171 \\
    \hline
    GRU & \multirow{2}{*}{0.8336$\pm$0.0027} & \multirow{2}{*}{\textcolor{olive}{\textbf{0.3096}}$\pm$0.0175} & \multirow{2}{*}{0.8357$\pm0.0015$} & \multirow{2}{*}{\textbf{0.6596}$\pm$0.0061} & 0.8688$\pm$0.0024 & 0.3188$\pm$0.0161 & 0.8401$\pm$0.0016 & 0.6759$\pm$0.0053 & 0.8691$\pm$0.004 & 0.3196$\pm$0.0208 & 0.8394$\pm$0.0019 & 0.6761$\pm$0.0073 \\
    GRU-LF & & & & & 0.8305$\pm$0.0108 & 0.3030$\pm$0.0144 &  0.8396$\pm$0.0014 & 0.6577$\pm$0.0071 & 0.8380$\pm$0.0137 & 0.3034$\pm$0.011 & \textcolor{red}{\textbf{0.8379}}$\pm$0.0016 & 0.6594$\pm$0.0047 \\
    \hline
    CNN & \multirow{2}{*}{\textcolor{orange}{\textbf{0.8201}}$\pm$0.0054} & \multirow{2}{*}{0.3289$\pm$0.0377} & \multirow{2}{*}{0.8448$\pm$0.0057} & \multirow{2}{*}{0.6646$\pm$0.0121} & 0.8556$\pm$0.0183 & 0.3095$\pm$0.0195 & \textcolor{red}{\textbf{0.8383}}$\pm$0.0065 & 0.6678$\pm$0.0109 & 0.8591$\pm$0.0056 &  0.4195$\pm$0.0933 & 0.8411$\pm$0.0055 & 0.7065$\pm$0.0307 \\
    CNN-LF & & & & & \textcolor{orange}{\textbf{0.8223}}$\pm$0.0081 & \textcolor{olive}{\textbf{0.2749}}$\pm$0.0151 & 0.8395$\pm$0.0052 & \textbf{0.6456}$\pm$0.0058 & \textcolor{orange}{\textbf{0.8193}}$\pm$0.0067 & \textcolor{olive}{\textbf{0.2827}}$\pm$0.0216 & 0.8483$\pm$0.0165 & \textbf{0.6501}$\pm$0.0086\\
\midrule
\end{tabular}
\end{adjustbox}
\label{tab:local_fn}
\end{table*}
From the results, it is evident that in all cases, the locally fine-tuned models are able to capture local characteristics further and result in lower prediction error. Considering the averaged NRMSE, we observe that both centralized and federated learning overcome the individual setting. Intuitively, this suggests that the globally-trained model captures similar dynamics from exogenous observations to that of the target base station and by performing local fine-tuning, we end up up with higher forecasting accuracy. We also emphasize that individual learning cannot provide generalization and hence, models trained under the centralized and federated settings are superior to the corresponding locally trained models. In fact, the fine-tuned CNN and GRU from both centralized and federated learning overcome the best model (GRU) on the individual setting. 

Taking a closer look at the resulting error per base station, we observe a similar behavior: in most cases, the fine-tuned models overcome the corresponding global models. Some exceptions occur on PobleSec, e.g., the CNN model under both centralized and federated learning. In ElBorn, the CNN model has the best performance, while the federated fine-tuned CNN slightly overcome the rest of the settings. In LesCorts, the fine-tuned centralized CNN slightly outperforms the corresponding federated model. Finally, in PobleSec, the behavior is different per learning setting since the best model architecture on one setting does not reflect the best model in the rest of the settings. This behavior is, again, attributed to arbitrary spikes and the presence of extreme outliers. 

Overall, the local fine-tuning step results in higher quality predictions and enables generalization for the centralized and federated settings as well. For instance, consider a generated model under either centralized or federated learning. The global model provides generalization among base stations since it captures global dynamics, while high quality forecasting can also be provided to an external, non-participating entity. In addition, an external base station can receive the trained model and perform local fine-tuning, which cannot be easily provided through the individual setting or when the distribution between two parties are highly skewed. Besides, federated learning enables a dynamic environment, where additional participants can emerge during training. This is a unique property of federated learning and cannot be modeled on the rest of the settings. Finally, federated training can occur on demand, e.g., once per day or week, to capture the dynamics of new observations in a timely manner and potentially lead to better predictions.

\subsection{Carbon Footprint Comparison}
In the previous sections, we identified the potential of centralized and federated learning for generalization. However, most trained models result in almost equivalent forecasting accuracy. For instance, we observed that the global federated GRU slightly overcomes the corresponding LSTM model regarding test forecasting error, but the latter provides more stability with respect to the standard deviation across runs on the validation set. 

In this subsection, we introduce another dimension to identify the top performing model and evaluate the considered models regarding energy consumption and carbon footprint ($CO_{2eq}$) using Carbontracker \cite{anthony2020carbontracker}. We also measure the total uplink and downlink consumption for the model weights transmission during federated training. 

Energy consumption and $CO_{2eq}$ emission are not directly comparable between centralized and federated learning. Centralized learning involves a third-party utilizing all available data and training a model for a specified number of epochs, while federated learning involves communication between the central server and participating entities per federated round. In Section \ref{learning_settings}, we stated that we selected a maximum 30 rounds with 3 local epochs during federated training and 270 epochs for the centralized setting to provide fair comparisons. Although this setting can provide generalized results regarding forecasting accuracy, measuring a model's efficiency is dependent on the number of observations, iterations (epochs) and the underlying hardware. Several works like \cite{liu2020traffic, Fekri2022Fl} directly compare federated rounds to centralized epochs. However, this approach does not reflect the actual number of operations involving participants. In our case, a single federated round corresponds to 3 participants, who perform 3 local epochs. In other words, considering the total number of iterations, participants perform 9 iterations on their data per federated round. Another major difference is that centralized training involves a single entity, while real-world federated training takes place concurrently on the selected participants. Since comparing centralized and federated learning regarding energy consumption is hard to achieve, we consider the following cases:
\begin{itemize}
    \item \textit{Centralized-1}: Centralized learning with as many iterations as the total number of iterations considered in the federated setting. That is, a total of 270 epochs since the considered federated setting involves 30 rounds, 3 local epochs and 3 participants.
    \item \textit{Centralized-2}: Centralized learning with epochs corresponding to federated rounds. This setting involves measurements considering 30 training epochs.
\end{itemize}
\begin{table}[b!]
\centering
\caption{Estimated energy (kWh) and carbon footprint (g) per setting.}
    \scriptsize
    \centering
    \begin{tabular}{c|c|c|c|c|c|c}
         Setting & Measure &  MLP & RNN & LSTM & GRU & CNN\\
         \hline
         \multirow{2}{*}{Centralized-1} & Energy & 0.0029 & 0.0032 & 0.0033 & 0.0033 & 0.0051\\
         & $CO_{2eq}$ & 0.8414 & 0.9415 & 0.9788 & 0.9803 & 1.4975\\
         
         \hline
         \multirow{2}{*}{Centralized-2} & Energy & 0.0003 & 0.0004 & 0.0004 & 0.0004 & 0.0006\\
         & $CO_{2eq}$ & 0.0914 & 0.1063 & 0.1064 & 0.1127 & 0.1685\\
         \hline
         \multirow{2}{*}{Federated} & Energy & 0.0011 & 0.0012 & 0.0014 & 0.0012 & 0.0019\\
         & $CO_{2eq}$ & 0.3216 & 0.3429 & 0.3975 & 0.3645 & 0.5640\\
         \midrule
    \end{tabular}
    \label{tab:energy_cen_fl}
\end{table}

Table \ref{tab:energy_cen_fl} shows the estimated energy consumption (kWh) and carbon footprint (g) per setting. Federated learning shows better results regarding both energy consumption and $CO_{2eq}$ emissions compared to centralized learning, when measured under the same conditions, i.e, the number of rounds data are accessed. That is mainly because of the smaller amount of data that a model has to iterate over on every client in contrast to a centralized counterpart that has to iterate over all data. It should be noted, that we do not consider the communication cost here, i.e., the environmental impact of data exchange between server and clients. However, we can easily assume that in a federated learning scenario this impact is much lower, since parties only exchange model parameters rather than whole sequences of data, as in a centralized scenario. Also, we should not ignore the fact that even a slight improvement in one base station for a specific amount of data, can be extremely beneficial if we examine the problem with many more base stations in a long period of time.

Taking these parameters into consideration, it can be deduced that federated learning not only allows for collaborative learning with high privacy guarantees, but can also help in minimizing the impact of running algorithms on the environment. 

Besides energy consumption, we are interested in the communication cost introduced on the server and the participants. To measure it, we select the federated round that achieved the lowest error regarding the validation set and measure the cost using the model size that needs to be transmitted from the server to participants and backwards. Note that each participant receives and transmits the model weights per federated round (i.e., client uplink equals to client downlink). The server transmits the model weights to the selected participants, which in our case, means that the model weights are being transmitted and received 3 times per federated round on the server side. Table \ref{tab:communication_cost} presents the minimum required transmissions of model weights to generate a global model. For each model, we report the corresponding size in KiloBytes (KB), the round that achieved the lowest validation error as well as the total uplink and downlink measurement at the client and server side in MegaBytes (MB). 

It is important to note here that we have refrained from attempting to manually transform these data transmissions into energy consumption and $CO_{2eq}$ emission estimates, since according to very recent results in~\cite{MYTTON20222032}, there are very few reliable data upon which to base our calculations.

Taking into account the amount of data transmitted, the most efficient model is RNN, but we are mainly interested in NRMSE score. Among LSTM and GRU, which are identified as the highest quality models, GRU results in more efficient computations. However, LSTM transmission size is relatively close to GRU and provides higher robustness. Considering that LSTM: i) performs the best in the validation set, ii) is the second best in the testing set, iii) it is robust against noise and iv) its efficiency is close to that of GRU, we select LSTM as our final solution. A critical future direction is to define a tunable measure that considers a model's error/accuracy, the convergence speed and its size to limit the quality-efficiency trade-off and identify the highest-performing model architecture. 

\begin{table}[b!]
\centering
\caption{Communication cost regarding the model size the federated round that achieved the lowest validation error.}
    \small
    \centering
    \begin{tabular}{c|c|c|c|c}
         Model & Size (KB) & Top Round & Client (MB) & Server (MB) \\
         \hline
         MLP & 563.2 & 15 & 8.4480 & 25.3440\\
         \hline
         RNN & 153.6 & 3 & 0.4608 & 1.3824\\
         \hline
         LSTM & 586.8 & 4 & 2.3472 & 7.0416 \\
         \hline
         GRU & 442.4 & 4 & 1.7696 &  5.3088 \\
         \hline
         CNN & 262.9 & 20 & 5.2580 & 15.7740\\
         \midrule
    \end{tabular}
    \label{tab:communication_cost}
\end{table}

\subsection{Aggregation Algorithms Comparison}
The previous experimental comparisons concerned federated learning under the FedAvg aggregation algorithm \cite{mcmahan2017fl}. To the best of our knowledge, most works in the generalized context of federated time-series forecasting, employ FedAvg without considering other aggregators \cite{taik2020load, liu2020traffic, liu2021anomaly, briggs2022fl, Fekri2022Fl}. Due to the heterogeneous nature of our task under four non-iid categories we are interested in evaluating the effectiveness of different aggregation algorithms. Besides the aggregation algorithms introduced in Section \ref{aggregators}, we also consider a simple averaging, denoted as \textit{SimpleAvg} and aggregation based on the median of received weights, denoted as \textit{MedianAvg}. The remaining considered aggregators aim to address the non-iid issue and introduce tunable parameters. For instance, FedProx \cite{Li2020fedprox} controls the weight divergence using a regularization parameter $\mu$. 
\begin{table}[b!]
    \small
    \centering
    \caption{Hyper-parameters per aggregator in the grid-search.}
    \begin{tabular}{c|c}
         Aggregator& Grid Parameters  \\
         \hline
         SimpleAvg & -\\
         \hline
         MedianAvg & -\\
         \hline
         FedAvg \cite{mcmahan2017fl} & -\\
         \hline
         FedProx \cite{Li2020fedprox} & $\mu \in \{10^{-3}, 10^{-2}, 10^{-1}, 10^{0}\}$\\
         \hline
         FedAvgM \cite{tzu2019fedavgm} & $\beta \in \{0., 0.7, 0.9, 0.97, 0.99, 0.997\}$\\
         \hline
         FedNova \cite{Wang2020fednova} & $\rho \in \{0, 10^{-3}, 10^{-2}, 10^{-1}, 0.99\}$\\
         \hline
         FedAdagrad \cite{reddi2021adaptive}& $\eta \in \{10^{-2}, 10^{-1}, 10^{0}\} $\\
         FedYogi \cite{reddi2021adaptive} & $\tau \in \{10^{-4}, 10^{-3}, 10^{-2}, 10^{-1}\}$\\
         FedAdam \cite{reddi2021adaptive}&  \\
         \hline
    \end{tabular}
    \label{tab:grids}
\end{table}

We select the federated LSTM model since it achieves high forecasting and efficiency performance. We train the model from scratch 20 times using different initialization seeds to obtain comprehensive results per aggregator. First, we tune the hyper-parameters introduced by the aggregation algorithms using a small grid search. Table \ref{tab:grids} reports the grids per aggregator. For FedAdagrad, we keep $\beta_1 = 0$; For FedYogi and FedAdam, we keep $\beta_1=0.9$ and $\beta_2=0.99$, according to \cite{reddi2021adaptive}. 
\begin{figure}[t!]
    \centering
    \includegraphics[width=\columnwidth]{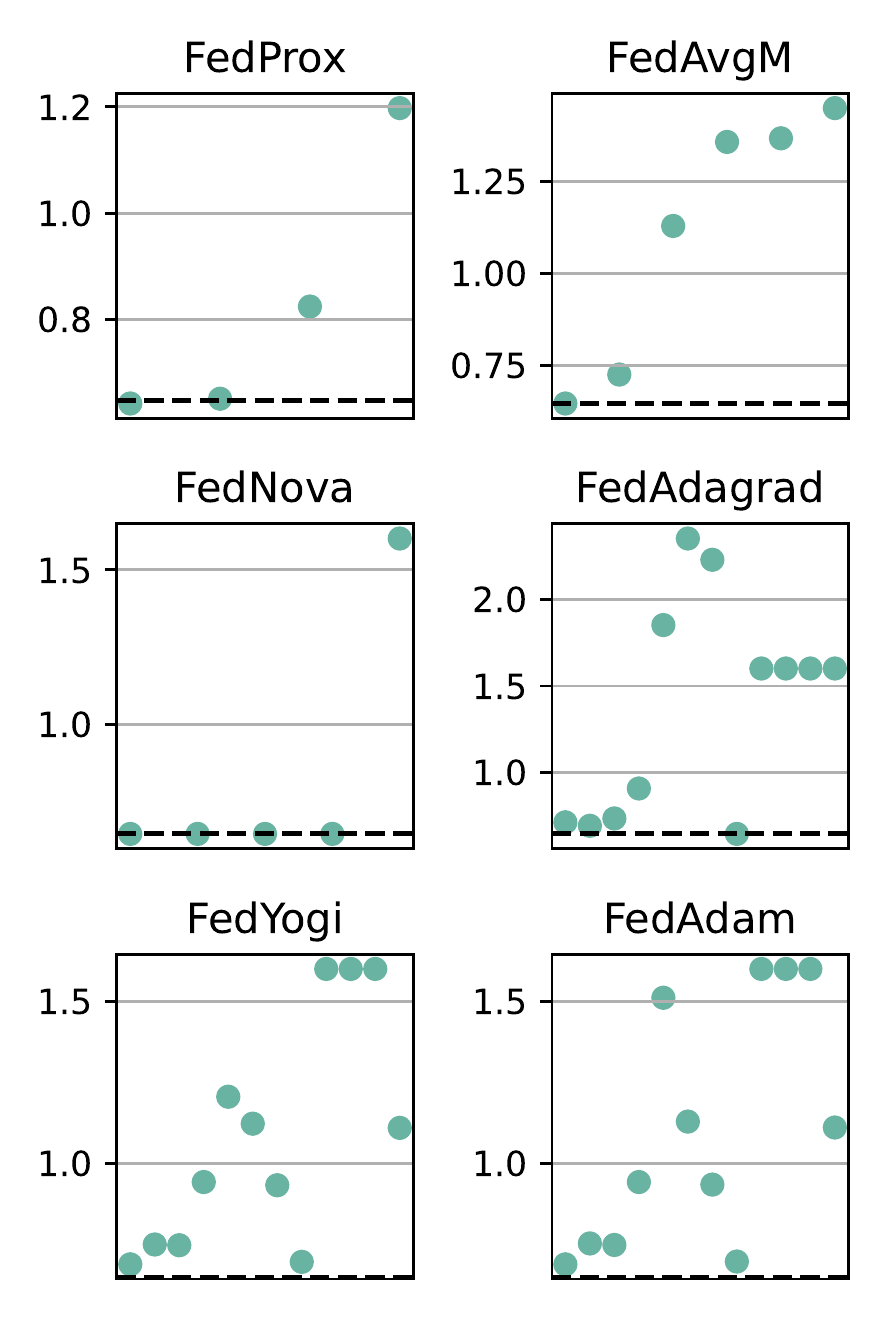}
    \caption{Averaged test NRMSE per grid and aggregator using the federated LSTM model. The dash line corresponds to the FedAvg baseline.}
    \label{fig:nrmse_val_lstm_aggregators_params}
\end{figure}

The resulting averaged NRMSE per grid in the test is illustrated on Fig. \ref{fig:nrmse_val_lstm_aggregators_params}. The dashed line corresponds to the FedAvg baseline and the points show the predictive error per grid. For \textit{FedProx}, we observe that the lower values of $\mu$, i.e., $\mu \in \{10^{-3}, 10^{-2}\}$, the better the predictive accuracy. Note that when $\mu = 0$, FedProx is equivalent to FedAvg. Since $\mu$ is a regularization parameter, higher values lead to slower convergence and hence, the selected 30 federated rounds do not lead to low prediction error. When $\mu$ is as low as $10^{-3}$, FedProx slightly overcomes the FedAvg baseline. For the \textit{FedAvgM} aggregator, we observe that the higher the value of $\beta$, the higher the error. Note that when $\beta=0$, FedAvgM is equivalent to FedAvg. From the experimental results, FedAvgM does not lead to lower prediction error and cannot outperform FedAvg. Regarding \textit{FedNova}, we observe that it follows the predictive accuracy of FedAvg when $\rho \leq 10^{-1}$ and slightly outperform FedAvg. \textit{FedAdagrad}, \textit{FedYogi} and \textit{FedAdam} require a pool of four hyper-parameters. In this work we optimize two of them, and the the rest is kept constant following their author observations \cite{reddi2021adaptive}. We observe that the predictive error of these aggregators is hugely affected by the underlying parameters and in many cases, they lead to high error values. FedAdagrad outperforms FedAvg only on a single case, while FedYogi and FedAdam do not lead to higher quality predictions than FedAvg. 

\begin{figure*}[ht!]
    \centering
    \includegraphics[width=0.85\textwidth]{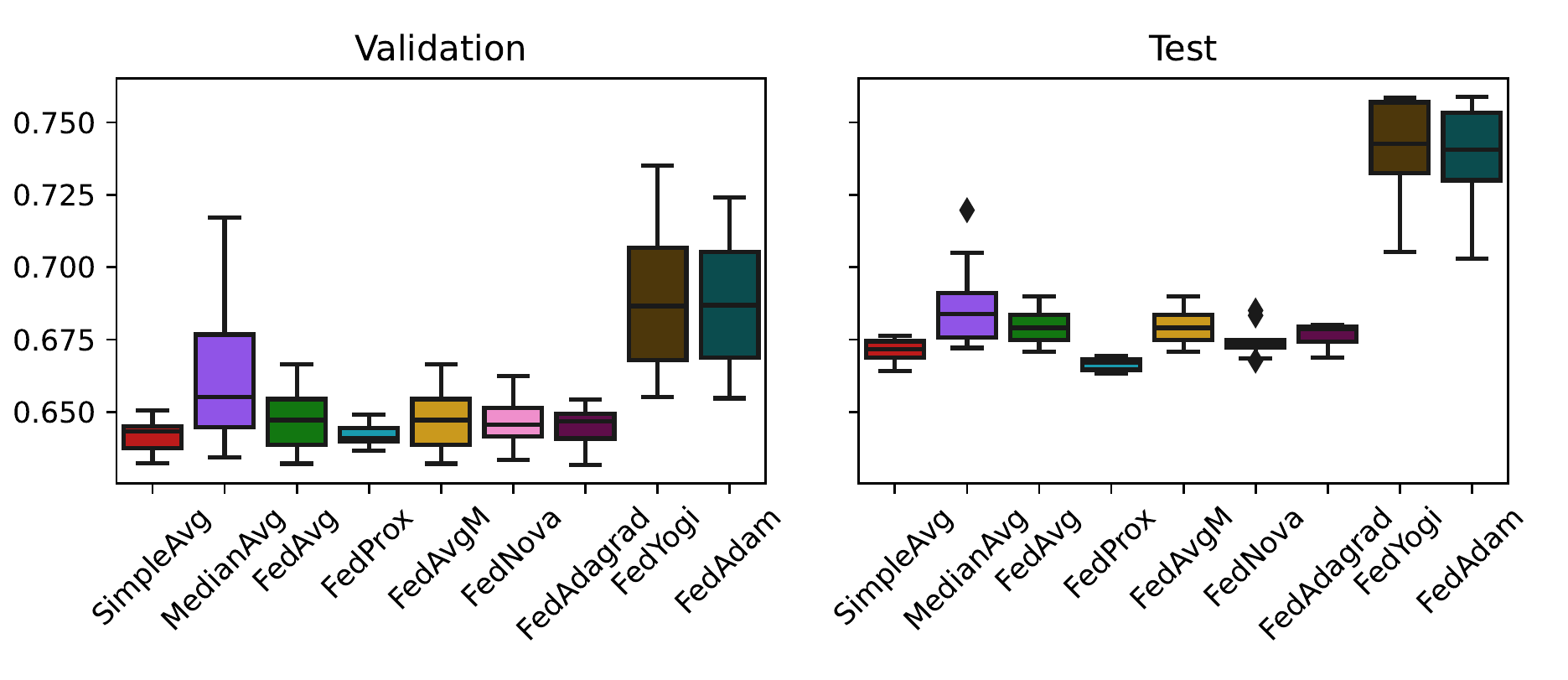}
    \caption{Validation and test averaged NRMSE considering the best hyper-parameters per aggregator using the federated LSTM model. The best hyper-parameters are selected based on the results on the validation set.}
    \label{fig:nrmse_val_test_lstm_aggregators}
\end{figure*}

Overall, it is evident that FedAvg is on par with FedProx and FedNova and state-of-the-art aggregation algorithms do not outperform the simple approach of FedAvg. In fact, FedOpt algorithms \cite{reddi2021adaptive} lead to unstable forecasting performance and the final global model is highly sensitive to the introduced hyper-parameters. The momentum parameter of FedAvgM does not improve the corresponding error, while FedProx and FedNova slightly outperform FedAvg in some cases. Fig. \ref{fig:nrmse_val_test_lstm_aggregators} shows the summarized NRMSE results per aggregation algorithm on the validation and testing sets. Note that we select the best hyper-parameters per aggregator based on the results on the validation set. Here, we consider two additional aggregators, i.e., SimpleAvg and MedianAvg. The NRMSE spread per aggregator clarifies the previous observations that FedProx and FedNova are on par with FedAvg. More precisely, \textit{FedProx} provides the highest robustness against random initialization and the lower averaged NRMSE on both the validation and testing sets. Next, despite its simplicity, SimpleAvg seems to outperform the rest of the aggregators both in NRMSE and stability. FedNova slightly overcomes FedAvg and shows relatively high stability. The rest of the aggregators show high spread, i.e., sensitivity to randomness, while FedAdagrad requires careful tuning of its parameters. We attribute this behavior to the unique patterns per base station, which exhibit a mixed type of skewness, thereby impeding the ability of more sophisticated aggregators to converge to a generalized model.

To the best of our knowledge there is only a single work \cite{li2022noniid} that systematically evaluates the behavior of different aggregators. However, Li et al. simulated non-iid data and do not use real-world datasets. In addition, the introduced scenarios in \cite{li2022noniid} only considered two cases of mixed non-iid data, namely, feature with label distribution skew and feature with quantity skew. In our case, there are four categories of non-iid data: the feature, target, quantity and temporal skew, a mixed case that is unique and directly applies to the real world. Nevertheless, our results are consistent to \cite{li2022noniid} and specifically:
\begin{itemize}
    \item There is no aggregator that systematically leads to higher quality models compared to FedAvg.
    \item FedProx is on par with FedAvg. 
    \item Federated learning is challenging with the presence of a mixed type of non-iid data.
\end{itemize}
A key difference in our results compared to \cite{li2022noniid} is that FedNova presents high stability. We attribute this behavior to the mixed type of non-iid data of the given dataset as well as to our use case. Nonetheless, there is still room for improvements and additional experimental studies, especially in the federated time-series forecasting research area, which have been given limited attention, should be carefully investigated.

\subsection{Scalability of Federated Learning}
\begin{figure}[t!]
    \centering
    \includegraphics[width=0.78\columnwidth]{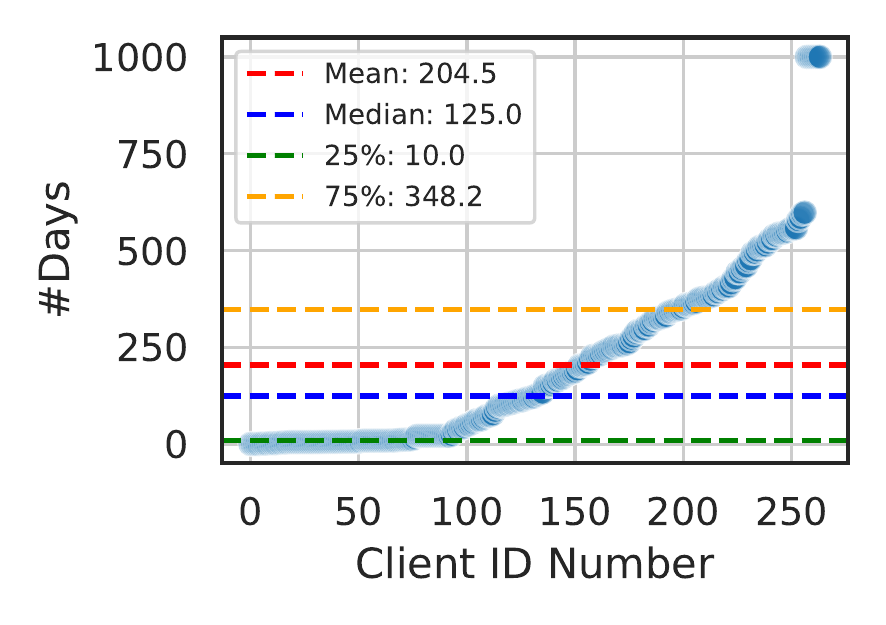}
    \caption{Data sizes per client in day interval. Each day corresponds to 720 observations.}
    \label{fig:synthetic_sizes}
\end{figure}

In this subsection, we introduce additional base stations to extend our use case beyond the existing three sites. To demonstrate the scalability of federated learning, we consider a scenario where multiple base stations own local data. In addition, we assume that all base stations are available and connected through a wired connection with the central server. All the experiments in this subsection are performed by training the LSTM model, which was previously identified as the top-performing model.

Due to the absence of publicly accessible traffic data for 5G and beyond networks, we employ generative models and train them using our up-to-date data. We train the generative models by considering either a single base station or exploring all possible combinations for our three base stations. Specifically, we trained the DoppelGANger \cite{Lin2021Gan} and CPAR \cite{zhang2022sequential} generative models. DoppelGANger focuses on generating multi-variate time-series using Generative Adversarial Networks (GANs) and is specifically designed for networking traffic flows. CPAR is a probabilistic auto-regressive GRU-based model for generalized time-series and stream flows generation.

\begin{figure}[t!]
    \centering
    \includegraphics[width=\columnwidth]{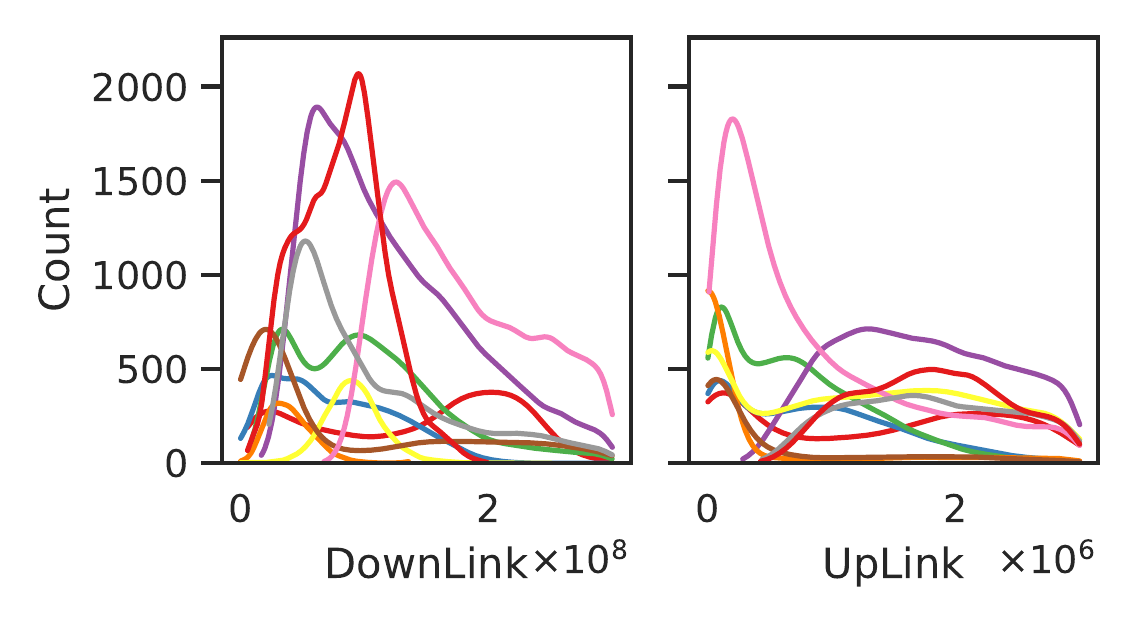}
    \caption{Distribution for the UpLink and DownLink values for 10 clients.}
    \label{fig:synthetic_distributions}
\end{figure}

Using the aforementioned models, we were able to generate a diverse dataset comprising 261 synthetic base stations, with varying number of samples and different distributions. Fig. \ref{fig:synthetic_sizes} illustrates the number of samples per client, including the three real-world base stations. Note that the number of samples are presented in day intervals, with each day consisting of 720 measurements. The number of days for which a synthetic client owns measurements varies from one day to 2.7 years, allowing us to model extreme quantity skew. On average, there are 204.5 days of measurements per client, with a median of 125 days. Besides data quantity skew, the synthetic base stations also exhibit distribution skew. Fig. \ref{fig:synthetic_distributions} presents an example of varying distributions among ten clients for the UpLink and DownLink measurements, including the original three base stations.

To demonstrate the impact of multiple clients with respect to scalability, we begin by examining the impact of data quantity on training time per (local) epoch. As mentioned previously, there are 264 base stations, with each one holding observations spanning from one day to 2.7 years. In federated learning, this variance in the number of observations results in significant differences in training times. In contrast, in a centralized scenario, the training time grows linearly with the total number of samples.

\begin{figure}[t!]
    \scriptsize
    \centering
    \includegraphics[width=\columnwidth]{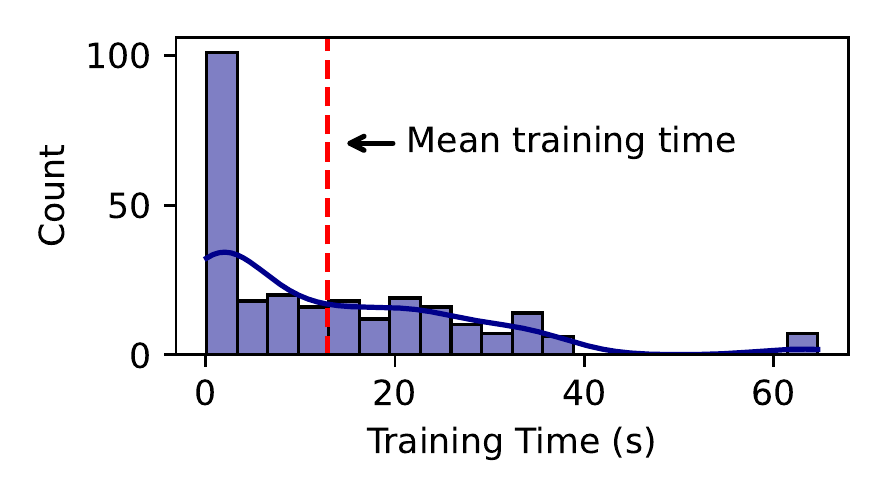}
    \caption{Federated training times distribution.}
    \label{fig:syn_training_times}
\end{figure}

Fig. \ref{fig:syn_training_times} shows the distribution of training times in the federated learning scenario for a single local epoch per client. As it can be observed, the training time ranges from as little as 1 second to 60 seconds. The former case corresponds to clients with limited data, e.g., one day of observations, while the latter case represents clients with observations spanning multiple years. The mean training time is 13 seconds per epoch, indicating that local training can be completed relatively fast. On clients with the larger amount of data, one local training epoch can be completed in approximately 1 minute. As such, considering the learning setting discussed in Section \ref{learning_settings}, where we selected three local epochs per participant, the training time can be as long as approximately 3 minutes. Thanks to the wired connection between the server and clients, the transfer time of model weights (which can be up to 600 KB, as shown in Table \ref{tab:communication_cost}) is negligible. In addition, federated learning enables parallel training execution, thus ensuring that a federated round would not exceed 4 minutes.

To quantify the training time in a centralized scenario, we partition the available base stations into bins. Specifically, we randomly shuffle the synthetic base stations and retain the first three IDs corresponding to the original real-world base stations. We train the same LSTM architecture considering a number of $\{3,$ $10, 25, 50, 100, 150, 200, 250, 264\}$ clients. Table \ref{tab:cen_times} reports the training time per considered number of clients, along with the number of total days in observations and the total data size in MB. Note that there is a significant increase between some client bins due to the inclusion of base stations that contains many observations. As expected, the training time grows with the number of clients and data size that is being considered. When we consider the total number of available clients, i.e., 264, the training time for a single epoch is about 1 hour, showing the superiority of federated learning, in which we identified that the maximum federated round completion takes about 4 minutes. Overall, in cases involving many clients, the training processing of federated learning outperforms the corresponding centralized setting due to its parallel execution. Hence, federated learning can scale much beyond centralized scenarios with respect to training times.

\begin{table}[t!]
\centering
\caption{Training times and data size to be transmitted in centralized learning per client bin.}
    \small
    \centering
    \begin{tabular}{c|c|c|c}
         \#Clients &\#Days &  Data Size (MB) &  Training Time (s)\\
         \hline
         3 & 37 &  2.68 & 2.4\\
         10 &87 &  6.27 & 5.7\\
         25 & 184 & 13.17 & 11.4\\
         50 & 464 & 33.14 & 29.6\\
         100 & 12630 & 901.99 & 892.1\\
         150 & 29784 & 2126.98 & 1880.2\\
         200 & 45790 & 3269.99 & 3059.3\\
         250 & 53906 & 3849.56 & 3487.1\\
         264 & 53994 & 3855.84 & 3512.8\\
         \midrule
    \end{tabular}
    \label{tab:cen_times}
\end{table}

\begin{figure}[t!]
    \centering
    \includegraphics[width=0.8\columnwidth]{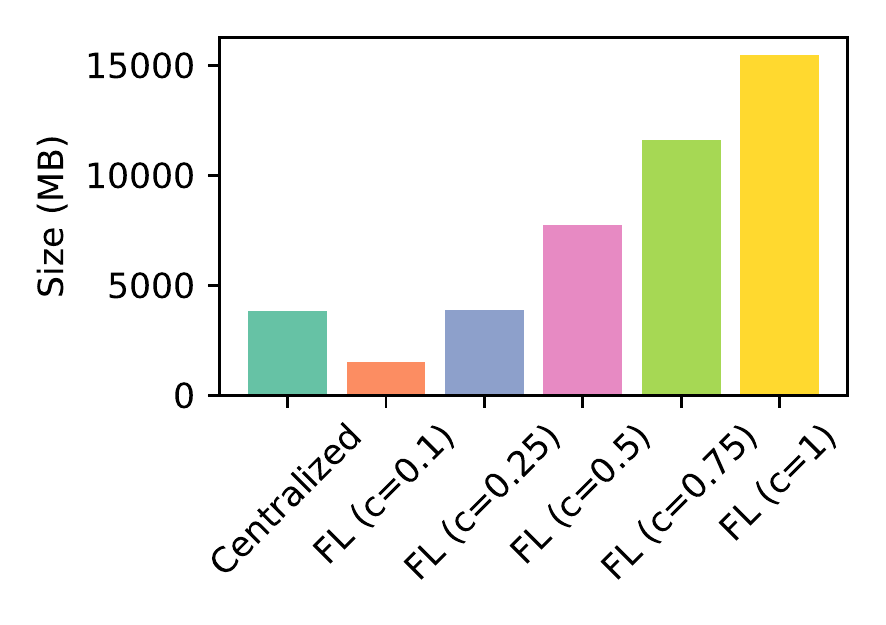}
    \caption{Size to be transmitted in centralized learning and federated learning scenarios.}
    \label{fig:com_size}
\end{figure}

Regarding communication costs, the data size to be transmitted in centralized setting is known since the available clients directly send their data to a central server. In federated learning, the communication cost, which is reflected in terms of model weights transfer, depends on the number of clients selected in each federated round to perform training, the model size and the number of federated rounds. To compare the two settings, we use the data size considering all clients. For the federated setting, we take into account the LSTM model (of size 586.8 KB), a constant of 50 federated rounds and a selection fraction from $\{0.1, 0.25, 0.5, 0.75, 1\}$ for our 264 base stations. In each federated round, the size to be transmitted is $2\times\text{model size}\times$ $\#\text{participants}$ since the model weights are broadcasted from the server to clients and from clients to the server. Fig. \ref{fig:com_size} illustrates the size to be transmitted in MB for both the centralized setting and the federated learning cases. It is observed, that when we choose a small fraction for selecting clients per federated round, i.e., 0.1 and 0.25, the total size that is transferred in federated learning is smaller or equivalent to that of the centralized setting. Specifically, the data size that needs to be transferred in a centralized scenario is about 3.9GB, while in federated learning and considering the fractions of 0.1 and 0.25, the total model weights transfer size is 1.6GB and 3.9GB, respectively. However, as we increase the selection fraction, federated learning requires a larger size to be transmitted. Thus, a selection of a small fraction is necessary for federated learning to scale beyond centralized settings is terms of communication costs.

\begin{figure}[t!]
    \centering
    \includegraphics[width=0.8\columnwidth]{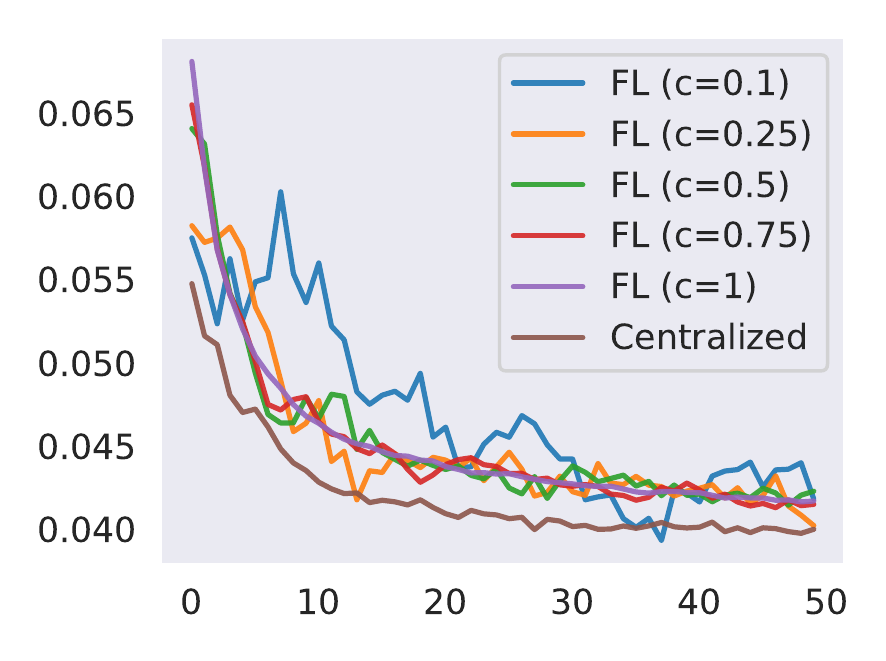}
    \caption{Training curves for different fraction selection in federated learning compared to centralized setting.}
    \label{fig:curves_fl_scenarios}
\end{figure}

To observe the trade-off between selection fraction and accuracy, we train the LSTM model using the aforementioned sampling fractions. We also train the centralized model, which is almost prohibitive in real-world scenarios due to the high computational costs and compare the achieved MAE on the validation set per iteration. Fig. \ref{fig:curves_fl_scenarios} shows the training curves for centralized and federated learning with client sampling. Here, we observe that centralized learning serves as the lower bound with respect to MAE. However, as previously mentioned, centralized learning is resource-intensive and cannot easily scale due to high computational demands and training time. On the other hand, for federated learning, we observe that in all selections fractions the model converges to nearly the same MAE, which is almost equivalent to the centralized setting. When we select a larger fraction (0.5 onwards), the model converges faster in the first epochs, but fails to achieve lower errors as federated rounds progress. By selecting a fraction of 0.25, the model converges almost as fast as the larger fractions, while achieving lower error compared to the federated settings with higher selection fractions. When we choose a fraction of 0.1, the model's convergence speed is the lowest among federated settings. Yet, it achieves even lower error than the centralized setting in round 38. Based on these observations, the optimal fraction should be low and in our case, lies between 0.1 and 0.25. Therefore, we conclude that for federated learning, a small selection fraction, along with intelligent client sampling, can achieve equivalent errors to the centralized setting, while offering faster training times, lower computational demands and lower communication costs than centralized learning. Ultimately, this enables federated learning to scale effectively across numerous clients, providing high predictive accuracy.

\section{Conclusion}
\label{sec:conclusion}
Generating high-quality traffic forecasting models with generalization ability is a challenging task considering the different date patterns in various base stations. In this work, we employed a federated learning approach to tackle several identified challenges owing to the non-iid data nature. Unlike centralized approaches, the proposed methods minimize business confidentiality and regulatory issues and federated learning paves the way for large-scale participation that can lead to building intelligent predictors. We conducted extensive experiments considering five deep learning architectures and compared three different learning settings. The results show that federated learning  has the potential for generalization, does not require the transmission of confidential data to a third-party and provides a dynamic execution environment. Furthermore, simulations conducted using the Carbontracker tool indicate that federated learning can yield substantial advantages in terms of reducing CO2 emissions and energy consumption compared to centralized learning, when assessed under identical conditions. In addition, local fine-tuning results in higher predictive accuracy and hence, federated learning could lead to more intelligent models than individual and centralized approaches. Finally, we showed that federated learning can scale up to numerous base stations, resulting into lower computational and communication costs than traditional machine learning.

A critical future direction is to focus on the \textit{preprocessing stage}, which according to our results, heavily affects the learning performance. Although preprocessing is an integral step in machine learning, there is limited conducted research on the influence of preprocessing applied to a federated setting. In addition, we showed that all learning architectures lead to equivalent forecasting accuracy. Based on our experimental study, LSTM and GRU lead to both lower errors and high robustness. Evaluating additional \textit{novel architectures}, such as promising transformers models, could result in higher-quality models. We also experimentally compared several different aggregation algorithms, some of which are specifically designed to tackle the non-iid data issues. However, we observed that in most cases, the aggregators are on par with the FedAvg baseline. Investigating \textit{additional aggregation algorithms} that can handle temporal dynamics could lead to higher predictive accuracy. We also showed that federated learning significantly reduces the training time and computational and communication costs compared to centralized learning. In the future, we will focus on further minimizing the associated costs using intelligent data and client sampling and model compression algorithms. Overall, we believe that our work sheds light on several unique challenges of time-series forecasting owing to the mixed skew type and can serve as a benchmark for various tasks including \textit{larger-scale federated learning}.

\section*{CRediT authorship contribution statement}
\textbf{Vasileios Perifanis:} Conceptualization, Methodology, Software, Formal analysis, Investigation, Writing - Original Draft, Writing - Review \& Editing, Visualization. \textbf{Nikolaos Pavlidis:} Methodology, Software, Validation, Writing - Review \& Editing, Visualization. \textbf{Remous-Aris Koutsiamanis:} Formal analysis, Validation, Writing - Review \& Editing, Supervision, Project administration. \textbf{Pavlos S. Efraimidis:} Conceptualization, Resources, Writing - Review \& Editing, Supervision, Project administration. 

\section*{Declaration of Competing Interest}
The authors declare that they have no known competing financial interests or personal relationships that could have appeared to influence the work reported in this paper.

\section*{Acknowledgments}
We thank the \href{https://supercom.cttc.es/}{SUPERCOM} Research Unit of CTTC for organizing the ML5G-PS-001 - "Federated Traffic Prediction for 5G and Beyond" challenge, part of the ITU AI/ML Challenge (2022) and specifically Dr. Francesc Wilhelmi for the constant support during the challenge.

\newpage
\appendix
\section{Supplementary Figures}
In this section, we present additional figures that supplement those described in the manuscript. Figure \ref{fig:input_series2} complements Fig. \ref{fig:series}, showing the remaining six input series. As it can be observed from these figures, there are significant variations in scales across variables. This variability is not only observed in the target series, which are also used as inputs (Fig. \ref{fig:series}), but also in the rest input series (Fig. \ref{fig:input_series2}). For example, the DownLink and UpLink measurements are in terms of $10^9$ and $10^8$ respectively, while other input series like MCS Up and Down Var are in terms of $10^{-7}$ and $10^{-8}$, respectively.
\setcounter{figure}{0}
\begin{figure}[htbp]
    \centering
    \includegraphics[height=0.95\textheight]{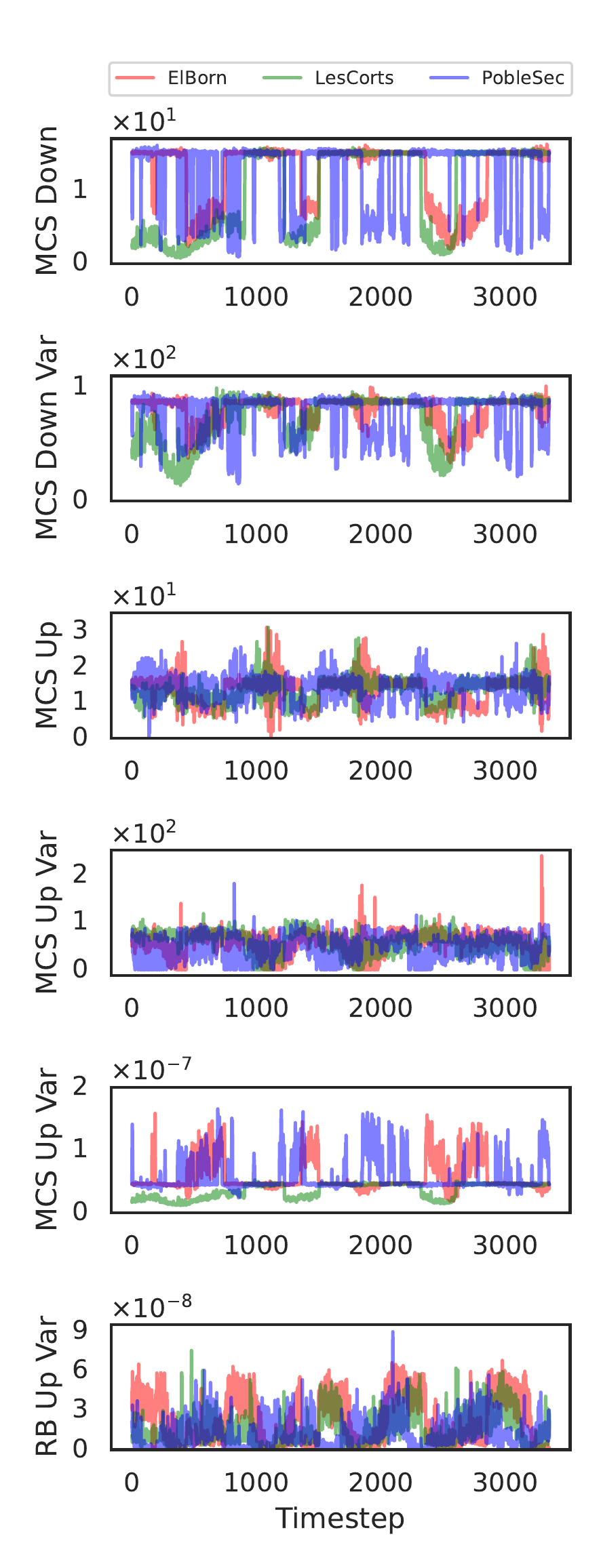}
    \caption{Additional input series per base station and timestep.}
    \label{fig:input_series2}
\end{figure}

In addition, Fig. \ref{fig:forecast2} completes the forecasts per base station using the federated LSTM model. In particular, we observe a similar behavior to the forecasts for the UpLink and DownLink values (Fig. \ref{fig:forecast}), where we discussed that although the model fails to capture the higher spikes in ELBorn and PobleSec, the overall patterns are well presented and federated learning leads to high predictive accuracy.
\setcounter{figure}{1}
\begin{figure*}[ht!]
    \centering
        \includegraphics[width=\textwidth]{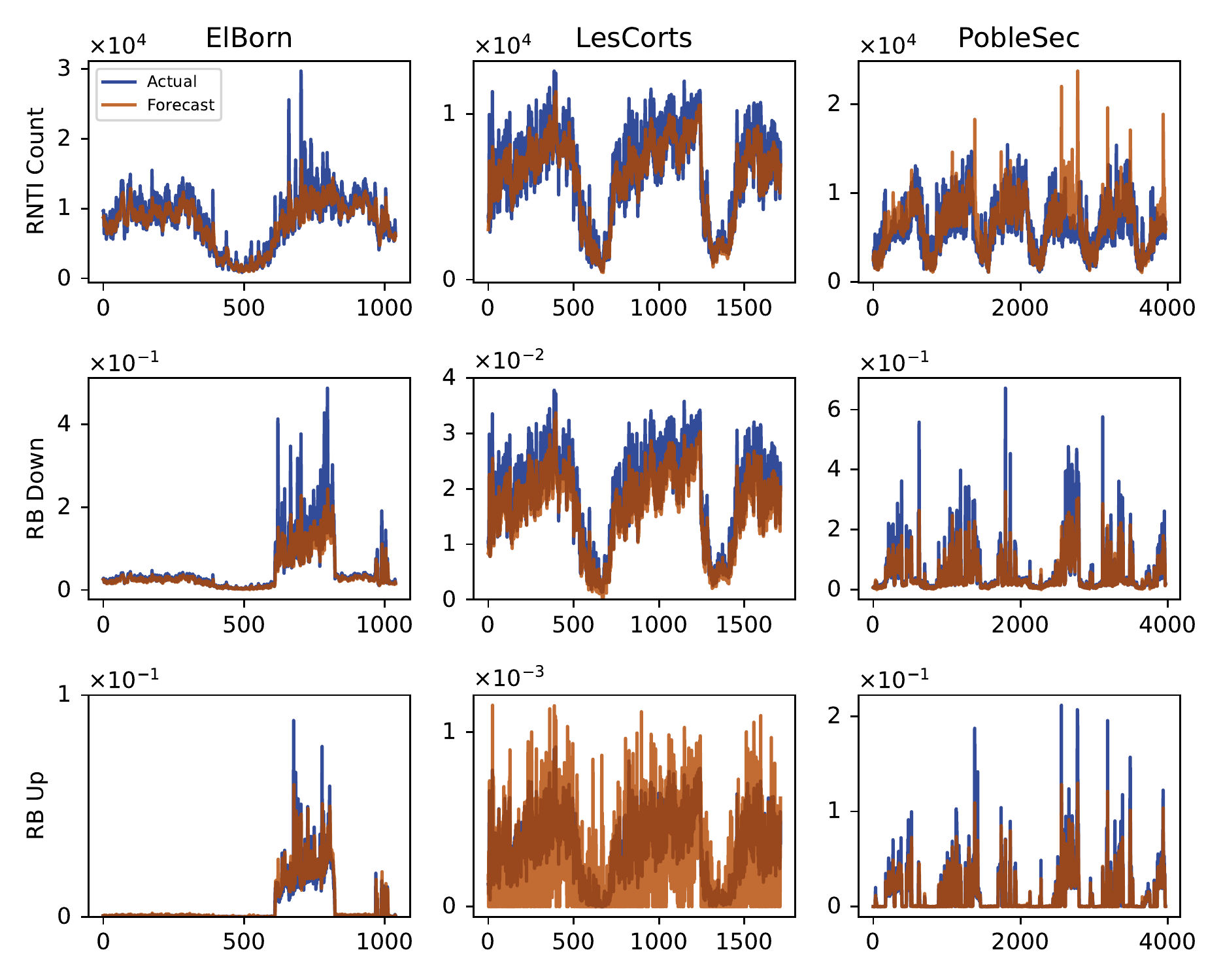}
    \caption{Additional predictions of the federated LSTM model against ground truth values per base station.}
    \label{fig:forecast2}
\end{figure*}

\bibliographystyle{elsarticle-num} 
\bibliography{cas-refs}

\subsection*{  } 
    \setlength\intextsep{0pt} 
    \begin{wrapfigure}{l}{0.13\textwidth}
        \centering
        \includegraphics[width=0.15\textwidth, height=0.15\textwidth]{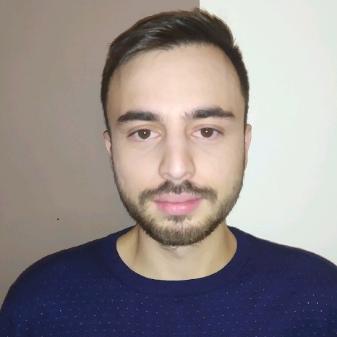}
    \end{wrapfigure}
    \noindent \textbf{Vasileios Perifanis} is a PhD Student at the Dept. of Electrical and Computer Engineering of the Democritus University of Thrace (Greece). He has a strong research interest in machine learning, with a focus on privacy-preserving federated learning and distributed computations.

\subsection*{  }
    \setlength\intextsep{0pt}
    \begin{wrapfigure}{l}{0.13\textwidth}
        \centering
        \includegraphics[width=0.15\textwidth, height=0.15\textwidth]{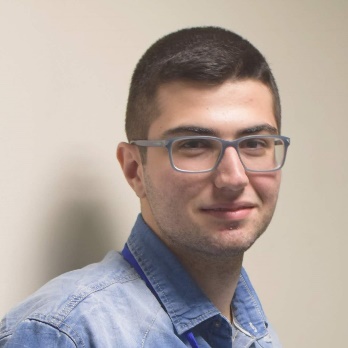}
    \end{wrapfigure}
    \noindent \textbf{Nikolaos Pavlidis} is a PhD student at the Dept. of Electrical and Computer Engineering of the Democritus University of Thrace (Greece). His main research interests are in federated learning and graph-based machine learning algorithms.

\subsection*{  } 
    \setlength\intextsep{0pt}
    \begin{wrapfigure}{l}{0.13\textwidth}
        \centering
        \includegraphics[width=0.15\textwidth, height=0.15\textwidth]{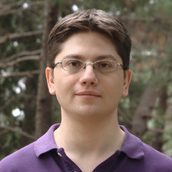}
    \end{wrapfigure}
    \noindent \textbf{Remous – Aris Koutsiamanis} is an Associate Professor at IMT Atlantique (Nantes, France) and a member of the STACK team of LS2N, a joint team with INRIA, focusing on the networking aspects of geo-distributed edge systems. Previously, he worked as a postdoctoral fellow at IMT Atlantique in the OCIF team of IRISA and focused on high performance network protocols for the Industrial Internet of Things (IIoT). He received his PhD from the Department of Electrical and Computer Engineering of the Democritus University of Thrace, Greece in February 2016 on the application of game theory to network QoS problems. His main research interests are in IoT, QoS and distributed resource management.
\subsection*{  } 
    \setlength\intextsep{0pt} 
    \begin{wrapfigure}{l}{0.13\textwidth}
        \centering
        \includegraphics[width=0.15\textwidth, height=0.15\textwidth]{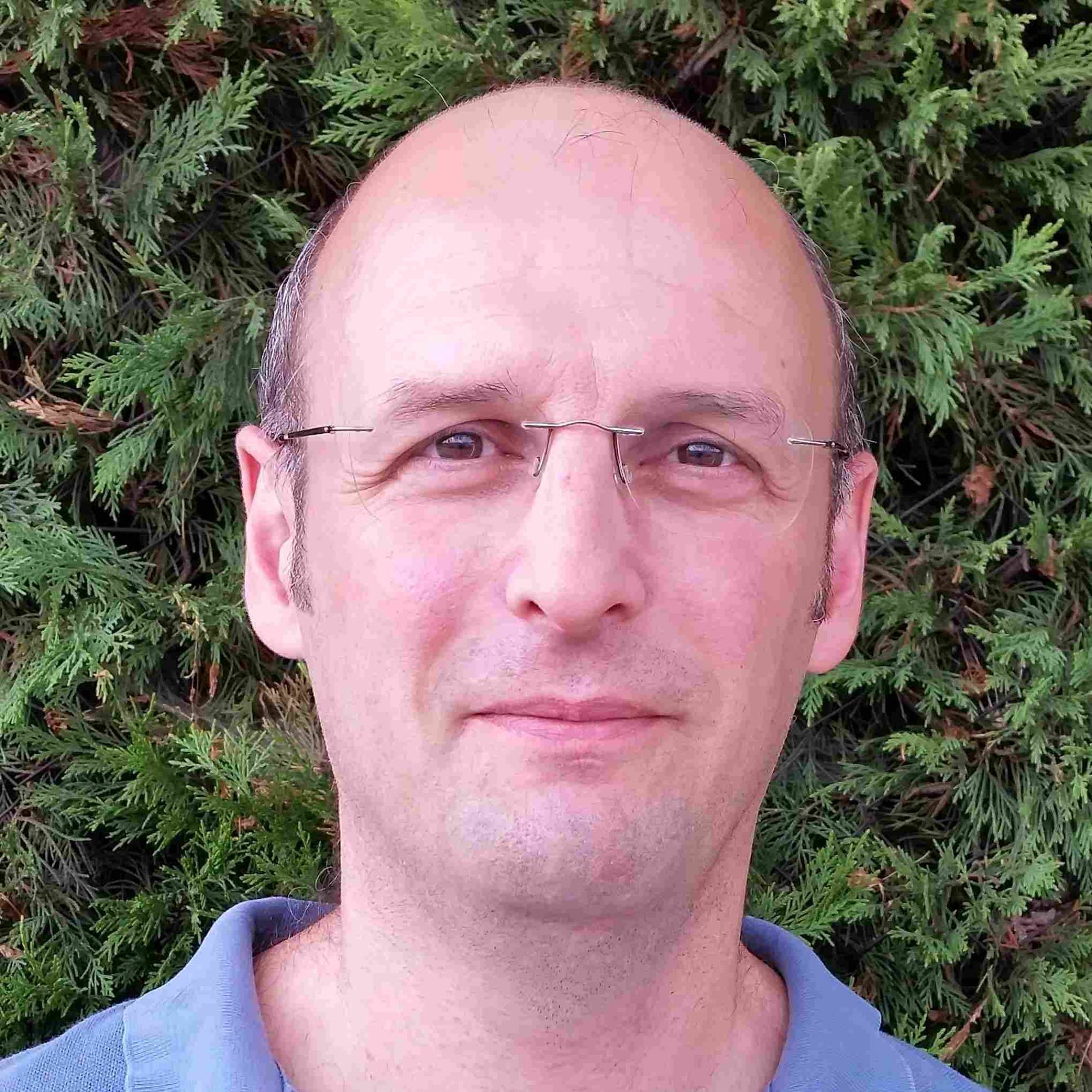}
    \end{wrapfigure}
    \noindent \textbf{Pavlos S. Efraimidis} is a Professor of Computer Science at the Dept. of Electrical and Computer Engineering of the Democritus University of Thrace (Greece) and an Adjunct Researcher of the Athena Research Center. He received his PhD in Informatics in 2000 from the University of Patras and the diploma of Computer Engineering and Informatics from the same university in 1995. His main work is on algorithms and his current research interests are in the fields of design and analysis of algorithms, graph theory and network analysis, federated machine learning, algorithmic game theory, and algorithmic aspects of privacy.

\end{document}